\documentclass[letterpaper]{article} 
\usepackage{aaai25} 

\usepackage{times}  
\usepackage{helvet}  
\usepackage{courier}  
\usepackage[hyphens]{url}  
\usepackage{graphicx} 
\usepackage{natbib}  
\usepackage{caption} 
\usepackage{algorithm}
\usepackage{algorithmic}

\usepackage{soul}
\usepackage{amsmath}
\usepackage{amsthm}
\usepackage{booktabs}
\usepackage{latexsym}

\usepackage{multirow}

\usepackage{amssymb}
\usepackage[T1]{fontenc}

\def\bi{\begin{itemize}}
	
	\def\ei{\end{itemize}}
\def\beq{\begin{equation}}
	\def\eeq#1{\label{#1}\end{equation}}
\def\ba{\begin{array}}
	\def\ea{\end{array}}
\def\i#1{\hbox{\it #1\/}}

\newcommand{\cred}{\color{red}}
\newcommand{\cmag}{\color{magenta}}

\newcommand{ \ccmk}{\checkmark$^*$}
\newcommand{ \dcmk}{\checkmark \textsuperscript{\textdagger}}
\newcommand{ \cmk}{\checkmark}
\newcommand{ \rx}{{\cred{$\times$}}}
\newcommand{ \rrx}{\cred{$\times^*$}}
\newcommand{\Del}{$\Delta$}

\def\bi{\begin{itemize}}
	
	\def\ei{\end{itemize}}
\def\beq{\begin{equation}}
	\def\eeq#1{\label{#1}\end{equation}}
\def\ba{\begin{array}}
	\def\ea{\end{array}}
\def\i#1{\hbox{\it #1\/}}

\long\def\BOC#1\EOC{\message{(Commented text )}}
\long\def\BOCC#1\EOCC{\message{(Commented text )}}
\long\def\BOCCC#1\EOCCC{\message{(Commented text )}}
\long\def\NB#1{\smallskip{[[\cmag {\bf N.B.} #1]]}\smallskip}

\long\def\NBB#1{}
\long\def\NB#1{}

\def\bi{\begin{itemize}}
	
	\def\ei{\end{itemize}}
\def\beq{\begin{equation}}
	\def\eeq#1{\label{#1}\end{equation}}
\def\ba{\begin{array}}
	\def\ea{\end{array}}
\def\i#1{\hbox{\it #1\/}}

\def\bcp{{\cal BC}+}

\usepackage{tcolorbox}
\tcbuselibrary{theorems,listings,skins,breakable}

\usepackage{newfloat}
\usepackage{listings}
\DeclareCaptionStyle{ruled}{labelfont=normalfont,labelsep=colon,strut=off} 
\floatstyle{ruled}
\newfloat{listing}{tb}{lst}{}
\floatname{listing}{Listing}

\def\bi{\begin{itemize}}
\def\ei{\end{itemize}}
\def\beq{\begin{equation}}
\def\eeq#1{\label{#1}\end{equation}}
\def\ba{\begin{array}}
\def\ea{\end{array}}
\def\i#1{\hbox{\it #1\/}}

\def\i#1{\hbox{\itshape #1\/}}

\def\iif{\hbox{\bf if}}

\def\causes{\hbox{\bf causes}}

\def\impossible{\hbox{\bf impossible}}
\def\nonexecutable{\hbox{\bf nonexecutable}}

\title{{\sf LLM+AL}: Bridging Large Language Models and Action Languages \\ for Complex Reasoning About Actions}

\author{
Adam Ishay\textsuperscript{\rm 1},
Joohyung Lee\textsuperscript{\rm 1,\rm 2} \\ 
}
\affiliations{
\textsuperscript{\rm 1}Arizona State University, AZ, USA\\
\textsuperscript{\rm 2}Samsung Research, S. Korea\\
\{aishay, joolee\}@asu.edu
}

\begin{document}

\maketitle   

\begin{abstract}	
Large Language Models (LLMs) have made significant strides in various intelligent tasks but still struggle with complex action reasoning tasks that require systematic search. To address this limitation, we propose a method that bridges the natural language understanding capabilities of LLMs with the symbolic reasoning strengths of action languages. Our approach, termed {\sf LLM+AL}, leverages the LLM's strengths in semantic parsing and commonsense knowledge generation alongside the action language's proficiency in automated reasoning based on encoded knowledge. We compare {\sf LLM+AL} against state-of-the-art LLMs, including {\sc ChatGPT-4}, {\sc Claude 3 Opus}, {\sc Gemini Ultra 1.0}, and {\sc o1-preview}, using benchmarks for complex reasoning about actions. Our findings indicate that, although all methods exhibit errors, {\sf LLM+AL}, with relatively minimal human corrections, consistently leads to correct answers, whereas standalone LLMs fail to improve even with human feedback. {\sf LLM+AL} also contributes to automated generation of action languages.
\end{abstract}

\section{Introduction} \label{sec:intro} 

Large Language Models (LLMs) have made significant strides in various intelligent tasks \citep{brohan2023can,kojima22large,huang23inner,zeng22socratic,yao24tree,besta24graph}, yet they often struggle with complex reasoning about actions, particularly in problems that demand systematic search. 
An emerging alternative is to use an LLM as a semantic parser to convert natural language into symbolic representations, such as Python programs  \citep{gao23pal,nye2021improving,olausson23linc}, Planning Domain Definition Language (PDDL) \citep{liu23llm+,guan23leveraging,xie23translating}, or logic programs \citep{ishay23logic}. These symbolic representations are then processed by dedicated symbolic reasoners. 

However, these methods have limitations. As demonstrated in this paper, for complex reasoning tasks, LLMs almost always fail to generate Python programs for searching for solutions, except in cases where the problem is a typical search task that LLMs may have memorized from the training corpus. Even in such instances, when small variations are introduced, LLMs struggle to adapt to the changes.

Using LLMs to generate PDDL could help address a broader range of action reasoning problems. However, PDDL is primarily designed as a standard language for task planning and is not well-suited for expressing more general forms of knowledge about actions, such as state constraints, indirect effects, default reasoning, and recursive definitions \cite{giu04}.

This paper presents a novel method that bridges the natural language understanding capabilities of LLMs with the symbolic reasoning strengths of action languages \citep{gel98,giu04}. Our approach, termed ``{\sf LLM+AL}," leverages the LLM's strengths in semantic parsing and commonsense knowledge generation alongside the action language's proficiency in automated reasoning about actions based on encoded knowledge.

Action languages are particularly well-suited for this purpose. They have an intuitive, natural language like syntax, feature formal semantics, and are supported by efficient computational tools. Action languages are designed for more knowledge-intensive reasoning than PDDL, encompassing not only task planning problems but also temporal prediction problems, which involve predicting what would happen if a sequence of actions is executed, and postdiction problems, where one infers the initial state given the current state and a past sequence of actions. Even when focused solely on task planning problems, action languages offer greater expressivity, such as representing indirect effects to address the ramification problem (e.g., the banana's location is determined by the monkey's location if the monkey is holding it; thus, any action that moves the monkey will indirectly affect the banana's location), and defaults (e.g., by default, a pendulum swings back and forth unless it is being held).
%
%
%
In particular, this paper leverages one of the latest members in this family, the action language ${\cal BC}$+ \citep{babb15action,babb15action1} due to its simplicity and expressivity, as well as the availability of the efficient $\bcp$ reasoner called  {\sc cplus2asp} \citep{babb13cplus2asp}.

The {\sf LLM+AL} pipeline leverages an LLM effectively across multiple stages, each serving a different purpose. First, given a reasoning problem in natural language, we use an LLM to generate a program signature and extract commonsense and domain-specific knowledge. Next, the LLM is tasked with converting this knowledge into $\bcp$ rules, guided by a prompt that details $\bcp$ syntax and semantics and is supplemented with several translation examples. After generating a complete $\bcp$ program, the LLM is tasked with doing a series of revisions if necessary, based on the output of the $\bcp$ reasoner. This pipeline facilitates the generation of correct $\bcp$ programs (with plans) or mostly correct $\bcp$ programs to users. Our code is available on  GitHub~\url{https://github.com/azreasoners/llm-al}, and the appendix can be found in the long version of the paper \cite{ishay25llm}.

Our findings indicate that {\sf LLM+AL}, even with basic descriptions of $\bcp$ and a few translation examples, is surprisingly adept at translating English into $\bcp$ rules and extracting relevant knowledge, thanks to the rich semantic understanding that LLMs acquired during pre-training. Despite occasional errors in knowledge extraction and rule generation---a challenge even for human experts---our approach demonstrates proficiency comparable to that of human experts, requiring only a few manual corrections to modify the erroneous $\bcp$ programs and produce correct solutions. 

Our focus is on problems that require deep rather than shallow reasoning, often involving systematic search, which LLMs typically struggle with. 
We evaluated our method using a benchmark proposed by \citet{mccarthy98elaboration}, which serves as a ``Drosophila" for assessing the elaboration tolerance of human-level AI---highlighting the importance of an AI's ability to represent and reason about new phenomena or altered circumstances.
The problem set includes several elaborations of the well-known Missionaries and Cannibals Puzzle (MCP), such as ``what if only one missionary and one cannibal can row?" and ``what if there are four missionaries and four cannibals instead?" We found that the most capable LLMs today, such as {\sc ChatGPT-4} \citep{chatGPT22}, {\sc Claude 3 Opus} \citep{claude24}, {\sc Gemini 1.0 Ultra} \citep{gemini23}, generally fail to produce  correct solutions for this benchmark, even after multiple iterations with human feedback pointing out errors in the answers. 
We also conducted experiments with {\sc o1-preview} \citep{openai24o1preview}, a novel type of LLM that utilizes test-time compute to handle more complex reasoning tasks.
While it outperformed other LLMs on the benchmarks, it still exhibited notable limitations.

In contrast, our approach, with relatively few human corrections, consistently leads to correct answers. This suggests that while current LLMs possess strong natural language understanding, they lack the systematic reasoning capability required to adapt effectively to new or altered scenarios. By integrating LLMs with action languages, our method demonstrates the potential for achieving a more robust and adaptable AI system.

\section{Preliminaries}  \label{sec:prelim}  

\subsection{LLMs for Planning} 

Several works have proposed applying LLMs to planning tasks \citep{huang22language, brohan2023can, huang23inner, singh23progprompt, yao23react}. For instance, SayCan \citep{brohan2023can} combines high-level actions with value functions, grounding LLMs in an environment. Inner Monologue \citep{huang23inner} integrates environmental feedback, including human feedback, into its pipeline, thereby enhancing robustness against agent errors. However, as noted in \citep{valmeekam22large, valmeekam23planning}, these methods struggle with more complex planning tasks.

One approach to address these limitations is using an LLM as an interface for symbolic reasoning engines. This includes generating executable Python code, as explored in recent work where natural language and Python program pairs are used to produce code for reasoning tasks \citep{olausson23linc, gao23pal, chen23program, lyu23faithful, singh23progprompt}. While this approach offloads much of the computation to the Python interpreter, it is not well-suited for planning tasks that involve constraints that are not easily expressible in a procedural language.

Another alternative is the use of Planning Domain Definition Language (PDDL) with LLMs. Some studies \citep{liu23llm+, xie23translating} have focused on translating English instructions into PDDL goals, assuming pre-existing PDDL action descriptions. Since only an instance file or goal needs to be generated, this setting is considerably simpler. Some recent works embrace a human-in-the-loop approach with LLMs, using human feedback when constructing domain models and executing plans \citep{guan23leveraging, huang23inner, yao23react}. 
Closest to our approach, \citet{guan23leveraging} employed an LLM for generating PDDL descriptions but noted that many manual corrections by PDDL experts were necessary due to errors in GPT-4’s translations, which impacted their execution by a PDDL solver. 

A number of recent works show some success using LLMs to iteratively revise their own output, surpassing baseline LLM performance while bypassing expensive human feedback \citep{madaan2024self, kim24language}. In particular, LLMs are well-suited for self-revision when they have access to external forms of feedback, such as external knowledge or tools (e.g., a code interpreter) \cite{kamoi24llms, stechly24self, guan23leveraging}. 

\subsection{Action Languages} \label{sec:action languages} 
\NBB{Need to rewrite for readability }

Action languages, such as ${\cal A}$ \citep{gel93a}, ${\cal B}$ \citep{gel98}, ${\cal C}$ \citep{giu98}, ${\cal C}$+ \citep{giu04}, ${\cal BC}$ \citep{lee13action}, and ${\cal BC}$+ \citep{babb15action,babb15action1}, represent subsets of natural language specifically designed for describing actions and their effects. These languages are often viewed as high-level notations of answer set programs \citep{lif08,brewka11answer,lee08a,gll06}, structured to effectively represent transition systems. {Key research topics in this field include the exploration of their expressive possibilities, such as indirect effects, triggered actions, defaults, and additive fluents \citep{giu04,gel98,lee03,inclezan16modular}. 
Such languages offer greater expressiveness than PDDL, which has been well-studied in the literature \citep{eyerich06golog,jiang19task}.} Despite the rich body of research surrounding action languages, a significant challenge remains: automation of action language generation, which we address in this paper. 

Constants in ${\cal BC}$+ are categorized into `fluent' and `action' constants. For instance, in the Blocks World domain, $\i{move}(B,L)$ represents an action constant with Boolean values (indicating whether the action is executed), while $\i{loc}(B)$ is a fluent constant with location values, where $B$ is a variable spanning over blocks.

The rules in ${\cal BC}$+ are called {\em causal laws}. An example is
\beq
\i{move}(B, L)\ \causes\ \i{loc}(B)=L
\eeq{r1}
which represents that moving a block $B$ to a location $L$ results in the block’s location being $L$. Another rule
\beq
\nonexecutable\ \i{move}(B,L)\ \iif\ \i{loc}(B_1) = B
\eeq{r2}
states that moving a block $B$ is not executable if another block $B_1$ is on top of it. Additionally,
\beq
\impossible\ \i{loc}(B_1)=B\land \i{loc}(B_2)= B
\eeq{r3}
$(B_1\ne B_2)$
illustrates a state constraint where two distinct blocks cannot occupy the same block.
The entire Blocks World domain can be described by these rules and a few extra ones. This succinctness is thanks to the separation between the representation language and the efficient constraint satisfaction algorithm for it. 
For a comprehensive review of ${\cal BC}$+, we refer the reader to~ \citep{lee13action,babb15action1}.

The input language of the ${\cal BC}$+ reasoner provides a convenient way of expressing ${\cal BC}$+ descriptions. It allows for declaring {\tt sort}s, {\tt object}s that belong to some {\tt sort}, and {\tt constant}s, such as fluents and actions.  
For example, the signature for the Blocks World is shown in Listing~\ref{lst:blocks_sign}
Additionally, the causal laws \eqref{r1}--\eqref{r3} above can be expressed in the language of the ${\cal BC}$+ as in Listing~\ref{lst:causal_laws}.

\begin{listing}
\caption{$\bcp$ signature for Blocks World}
\label{lst:blocks_sign}
\begin{lstlisting}
:- sorts
   loc >> block.

:- objects
   b1, b2, b3, b4     :: block;
   table              :: loc.
   
:- constants
   loc(block)        :: inertialFluent(loc);
   move(block, loc)  :: exogenousAction.
\end{lstlisting}

\end{listing}

\begin{listing}
\caption{$\bcp$ rules for Blocks World ({\tt \%} is for comments)}
\label{lst:causal_laws}
\begin{lstlisting}
% Moving a block changes its location.
move(B,L) causes loc(B)=L. 

% Can't move block with something on it.
nonexecutable move(B,L) if loc(B1)=B.

% Two blocks cannot be on the same block.
impossible loc(B1)=B & loc(B2)=B & B1\=B2. 
\end{lstlisting}
\end{listing}

\section{Our Method} \label{sec:methodology} 

\begin{figure*}[ht]
\begin{center}
\includegraphics[width=1.9\columnwidth,height=5.4cm]{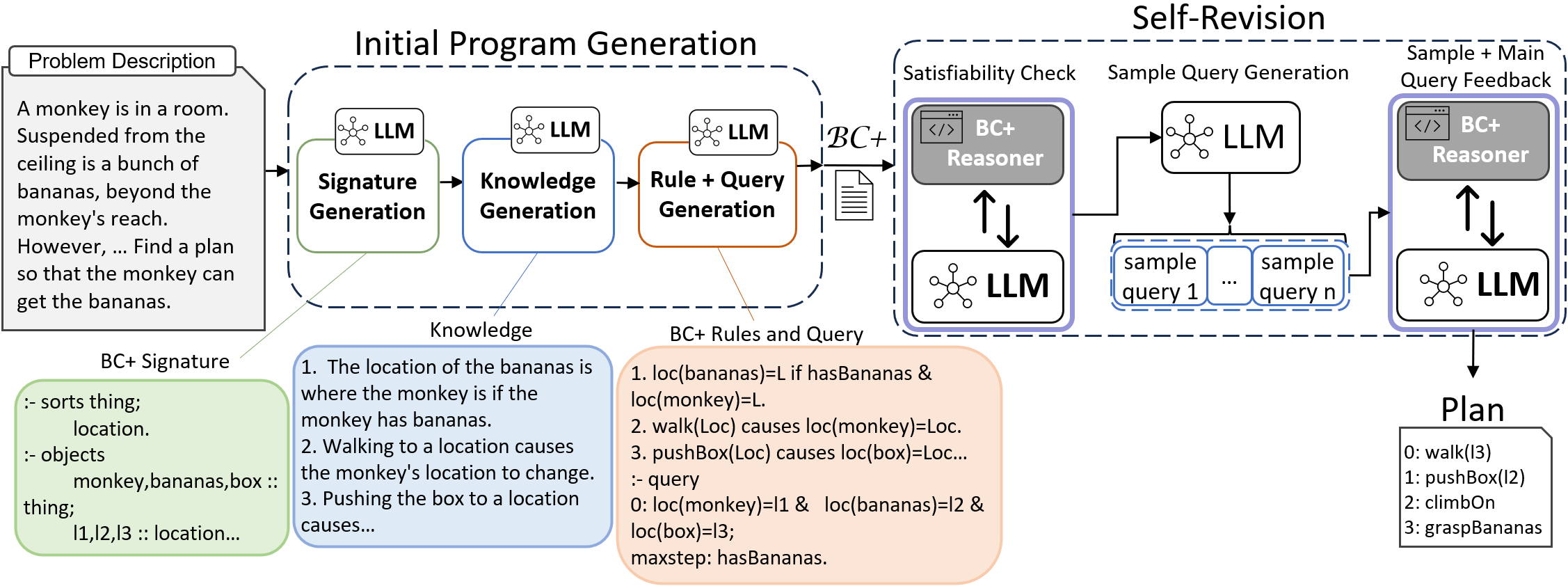}
\end{center}
\caption{{\sf LLM+AL} pipeline}
\label{fig:main_fig}
\end{figure*}

Our framework, as depicted in Figure~\ref{fig:main_fig}, comprises four principal components: $\bcp$ Signature Generation, English Knowledge Generation, $\bcp$ Rule and Query Generation, and Self-Revision. The $\bcp$ Signature Generation is responsible for defining necessary symbols. English Knowledge Generation involves extracting and structuring relevant information from the natural language problem description,  while $\bcp$ Rule Generation focuses on translating this structured knowledge into formal $\bcp$ rules, thereby bridging the gap between natural language understanding and symbolic reasoning. Finally, Self-Revision iteratively refines the $\bcp$ signature, rules, and query, with feedbacks from the $\bcp$ reasoner when run on a set of queries generated by the LLM. The result is either a correct program (and a correct solution) or a program that requires a typically small number of corrections to yield a correct solution.  



\subsection{Input}  \label{ssec:input}  

The input is a natural language description of the problem, including descriptions of types, objects, and actions involving them, along with a query in natural language.  
For example, the input for the Missionaries and Cannibals puzzle (MCP) is given in Appendix~A.1. 


\subsection{$\bcp$ Signature Generation} 

Given the problem description in English, this step generates a signature in ${\cal BC}$+ syntax. Writing such a $\bcp$ program typically starts with understanding the problem and considering the dynamics (knowledge) required, thinking about what fluent and action constants are useful, and then writing rules about them. We present the LLM with the problem and prompt it to generate important parts of the problem in natural language before signature generation. 
The Signature Generation prompt (See Appendix B.1) contains an introduction to $\bcp$ and a few example translations of the English description to important knowledge, an analysis of constants and their natural language reading, and finally, a ${\cal BC}$+ signature. Only the natural language reading of constants and signature are passed to the rest of the pipeline.
For the MCP puzzle, the generated signature can be found in   Appendix~A.2.

\subsection{Knowledge Generation} \label{ssec:k-extraction} 

We leverage an LLM to extract relevant knowledge from a problem description for use by the ${\cal BC}$+ reasoner.
This is achieved by using a prompt that includes instructions and a few example domain descriptions in English, along with their corresponding knowledge statements in English.
The generated knowledge statements broadly fall into two categories, {\em commonsense knowledge} and {\em domain-specific knowledge}. Commonsense knowledge refers to information that is not explicitly stated in the problem description and usually takes the form of cause-and-effect relationships. For example, in the MCP domain, this step correctly generates the commonsense knowledge statements: \textit{``crossing a vessel causes the location of the vessel to change''} and \textit{``crossing a vessel causes the number of a group at a location to decrease by the amount of members on the vessel.''} 
Enumerating such commonsense knowledge can be tedious and prone to omissions. Therefore, we find it beneficial to use an LLM for this task.  
Domain-specific knowledge is directly tied to the given information about the problem, for instance,
\textit{``Missionaries should not be outnumbered by cannibals, or they will be eaten."} 
The full prompt for extracting knowledge is shown in Appendix~B.2. 
The complete knowledge generated for MCP is shown in Appendix~A.3. 


\subsection{Rules and Query Generation} \label{sec:knowledge correction} 

Further leveraging the capabilities of LLMs, we use them to convert natural language into a symbolic representation. 
This is done with a prompt that contains a brief introduction of action language $\bcp$ and few-shot examples of natural language knowledge translated into ${\cal BC}$+ rules. 
This prompt, together with the natural language reading of constants, signatures, and the English knowledge generated in previous steps, is given as input to the LLM, which then generates $\bcp$ causal laws. The full prompt is given in Appendix~B.3.

For example, the knowledge \textit{``crossing a vessel causes its location to change"} is turned into:

\begin{small}
\begin{verbatim}
cross(V) causes loc(V)=Loc if going(V)=Loc.
\end{verbatim}
\end{small}


\subsection{Self-Revision} \label{sec:self-revision} 

The programs generated so far may have errors, either syntactically or semantically. This step leverages an LLM to revise the generated $\bcp$ program based on the reasoner's output. 

\smallskip
\noindent
\textbf{Satisfiability Check.}\ \ \  
First, without considering any initial or goal state, the pipeline performs a {\em satisfiability check}, which runs the $\bcp$ reasoner to ensure if the generated $\bcp$ signature and rules (without a query) are satisfiable. Typical errors detected in this step are syntax errors and reported by the $\bcp$ reasoner. Based on the output message of the $\bcp$ reasoner (e.g., syntax errors on some rules or undeclared constants), the LLM is prompted to update the signature and/or rules accordingly. This step repeats until either the program is satisfiable, in which case the pipeline proceeds to Sample Query Generation, or the maximum number of allowed revisions is reached, in which case the pipeline ends on this step, settling on the current signature, rules, and the main query.

\smallskip\noindent
\textbf{Sample Query Generation.}\ \ \ 
Next, the pipeline prompts the LLM to generate a small set of simple sample queries. These queries will later serve to check that the program correctly implements domain-specific rules and constraints (e.g., actions lead to expected changes in the state, preconditions for actions are respected, etc.). The LLM is instructed to append either ``(satisfiable)'' or ``(unsatisfiable)'', depending on whether the query is expected to be satisfiable or unsatisfiable based on the LLM's discretion. For example, a sample query for the Missionaries and Cannibals puzzle could involve an action which is is expected to be disallowed, such as crossing on the boat with more than the allowed capacity, which would be appended with ``(unsatisfiable)''.

\smallskip
\noindent
\textbf{Sample and Main Query Feedback.}\ \ \
The pipeline automatically executes each sample query and the main query using the $\bcp$ reasoner, and the outputs are provided to the LLM. The LLM is then tasked with verifying that the outputs align with the domain and revising the $\bcp$ signature, causal laws, main query, and/or sample queries as needed. If no changes are required or the maximum number of revisions is reached, the resulting $\bcp$ program and the $\bcp$ reasoner’s output are finalized. Otherwise, the process repeats.
See Appendix~A.6. 
for an example of the Self-Revision step.

\section{Experiments} \label{sec:experiments} 



\noindent
\textbf{Benchmarks.}\ \ \  We consider benchmarks focusing on complex reasoning. The first set is from \citep{mccarthy98elaboration}, where McCarthy proposed several variations of the well-known Missionaries and Cannibals puzzle. The second set consists of several well-known puzzles along with our own variations.
 
\smallskip
\noindent
\textbf{Baselines.}\ \ \  For the baseline LLMs, we use  {\sc ChatGPT-4}, {\sc Claude 3 Opus}, {\sc Gemini 1.0 Ultra}, and {\sc o1-preview}.\footnote{All baseline experiments were conducted in May 2024, except for {\sc o1-preview}, which was conducted in September 2024. For other experiments with more recent LLM, please refer to our github page.} These baseline models are provided with problem descriptions in natural language, as outlined in Section~\ref{ssec:input}, and tasked with finding a solution.
Additionally, we evaluate {\sc ChatGPT-4} using its code interpreter feature to generate Python programs for solving the problems. This method is referred to as {\sc ChatGPT-4+Code}. Similar to our pipeline, we prompt {\sc ChatGPT-4+Code} to iteratively revise its generated programs as many times as needed.

Recent studies have shown that methods like Chain-of-Thought (CoT) \citep{wei22chain}, Program-of-Thought (PoT) \citep{chen23program}, and similar prompting techniques applied to enterprise LLMs have not been effective and may even perform worse \citep{valmeekam23planning, chen23you, stechly24chain}. This is likely because such methods are already incorporated during the instruction following training of these LLMs. In preliminary experiments, we observed the same, so we did not include these methods in the final experiments.
\subsection{Experiment Results} \label{sec:mcp} 

\begin{table*}[h]\setlength{\tabcolsep}{2.5pt}
\small
\renewcommand{\arraystretch}{.8}
 \centering 

\begin{tabular}{|l||c||c|c|c|c|c|c|c|c|}
\hline
Problem                                             &  Opt.   & Chat   & Claude   &Gemini  & ChatGPT4  & o1         & LLM    \\     
                                                    &  Length & GPT4   & 3 Opus   & 1 Ultra &    Code  & (preview)  &  +AL   \\
\hline
\hline                                   
MCP (basic)                                       &  11     &  \rx   &  \rx     &  \rx   &\cmk [1]  & \cmk  &   \cmk         \\ 
1 (the boat is a rowboat)                         &  11     &  \rx   &  \rx     &  \rx   &\cmk [1]  & \cmk  &   \cmk         \\
2 (missionaries and cannibals can exchange hats)  &  11     &  \cmk  &  \rx     &  \rx   & \cmk [2] & \rx   &   $\triangle$(1)   \\
3 (there are 4 missionaries and 4 cannibals)      & (unsolvable) &  \rx   &  \rx     &  \rx   & \Del [6] & \cmk  &   $\triangle$(2)   \\
4 (the boat can carry three, 5 missionaries/cannibals)                      & (unsolvable) &  \rx   &  \rx     &  \rx   &  \rx [2] & \cmk  &   \cmk          \\
5 (an oar on each bank)                           &  13     &  \rx   &  \rx     &   \rx  &  \rx [1] &  \rx  &   $\triangle$(2)   \\
6 (only one missionary and one cannibal can row)  &  13     &  \rx   &    \rx   &   \rx  &  \rx [1] & \rx   &   $\triangle$(4)   \\
7 (missionaries cannot row)                       & (unsolvable) &  \rx   &   \rx    &   \rx  & \Del [6] & \cmk  &   \cmk          \\
8 (a very big cannibal must cross alone)          &  15     &  \rx   &   \rx    &  \rx   & \rx  [2] & \rx   &   $\triangle$(2)   \\
9 (big cannibal and small missionary)             &  11     &  \rx   &   \rx    &   \rx  & \rx  [1] & \rx   &   $\triangle$(6)   \\
10 (a missionary can walk on water)               &  7      &  \rx   &   \rx    &  \rx   & \dcmk [1]& \rx   &   $\triangle$(7)   \\
11 (missionaries can convert cannibals)           &  9      & \rx    &  \rx     &  \rx   &\dcmk [2] & \rx   &   \dcmk         \\
13 (there is bridge that can cross two, 5 missionaries/cannibals) &  4      &  \rx   &   \rx    &  \rx   & \rx  [6] & \dcmk &   \dcmk         \\
14 (the boat leaks with two people on it)         &  11     & \cmk   &  \rx     &  \rx   &  \rx [1] & \rx   &    $\triangle$(3)  \\
16 (there is an island, 5 missionaries/cannibals)                           &  19     & \rx    &  \rx     &  \rx   &  \rx [4] & \rx   &   \cmk         \\
17 (cannibals can become hungry, 4 missionaries/cannibals)                  &  13     & \rx    &  \rx     &  \rx   &  \rx [4] & \rx   &   $\triangle$(3)   \\
19 (there are two sets of groups)                 &  22     & \cmk   & \rx      & \rx    &  \rx [1] & \cmk  &    $\triangle$(1)  \\
\hline

Total                                              &         & 3       &   0      &   0    &   5+2$\triangle$   &  7  & 7+10$\triangle$   \\
\hline
\end{tabular}
\caption{Performance on MCP and its elaborations. (We exclude a few elaborations because they require probabilistic reasoning or the descriptions are vague.)  $\Delta$ indicates human intervention was used to produce a correct result.
 [$n$] indicates the number of attempts {\sc ChatGPT-4+Code} makes at writing a program, and ($n$) indicates the number of manual corrections required for {\sf LLM+AL}. \textsuperscript{\textdagger} indicates that the solution found was not optimal. All elaborations are listed in Appendix~E. \iffalse \ref{sec:mcp_elaborations}.\fi } 
 \label{tb:mcp_table}
\end{table*}

\begin{table*}\setlength{\tabcolsep}{2.5pt}
\small
\renewcommand{\arraystretch}{.8}
 \centering 

\begin{tabular}{|l||c||c|c|c|c|c|c|c|c|}
\hline
Problem                    &Opt.      &  Chat    & Claude   &Gem. 1.0& {\sc ChatGPT-4}& o1 & LLM+  \\ 
                           &Length    &   GPT4   & 3 Opus   &Ultra   & +Code          & preview &   AL   \\
\hline
\hline                           
River Cross (basic)        &  7       & \cmk     &   \cmk    & \cmk & \cmk [5] &  \cmk &\cmk        \\ 
River Cross  (var1)        &  6       & \rx      &    \rx    &  \rx & \rx  [3] &  \cmk &\cmk        \\ 
Tower of Hanoi (3-disk, basic)        &  7       & \rx      &   \rx     & \rx  & \cmk [1] & \cmk  &\cmk        \\ 
Tower of Hanoi (3-disk, var1)         &  6       & \rx      &   \rx     & \rx  & \rx  [1] & \cmk  & $\triangle$(1) \\ 
Tower of Hanoi (5-disk, basic)        &  31      & \ccmk    &   \rx     & \rx  & \cmk [1] & \cmk  & \cmk        \\ 
Tower of Hanoi (5-disk, var1)         &  27      & \rx      &   \rx     & \rx  & \rx  [4] & \rx   & \cmk        \\ 
Tower of Hanoi (7-disk, basic)        &  127     & \rx      &   \rx     & \rx  & \cmk [1] &  \rx  & \cmk        \\ 
Tower of Hanoi (7-disk, var1)         &  11      & \rx      &   \rx     & \rx  &  \rx [2] &  \rx  & $\triangle$(1)  \\ 
Sudoku1                    &  0       & \ccmk    &   \rx     & \rx  & \cmk [1] &  \rx  & $\triangle$(2) \\ 
Sudoku2                    &  0       & \ccmk    &   \rx     & \rx  & \cmk [1] &  \rx  & \cmk        \\ 
Sudoku3                    &  0       & \rrx     &   \rx     & \rx  & \cmk [3] &  \rx  & \cmk        \\ 
Sudoku (var1)              &  (unsolvable) &  \ccmk   &   \rx     &  \rx & \Del [3] &  \rx  & $\triangle$(1) \\ 
Sudoku (var2)              &  (unsolvable) &  \rrx    &   \rx     &  \rx & \cmk [0] &  \rx  & $\triangle$(6) \\ 

\hline                        

Total                      &         &    5      &   1       &  1    &  8+1\Del  & 5  &  8+5\Del    \\ \hline 
\end{tabular}
\caption{Performance on some puzzles and their variations.
 $^*$ indicates that the LLM voluntarily generated Python code.  
}
\label{tb:cross_table}
\end{table*}

\smallskip\noindent
\textbf{Benchmark Performance.}\ \ \  
As shown in Table~\ref{tb:mcp_table}, the baseline LLMs perform poorly on the MCP elaboration problems. Neither {\sc Claude 3 Opus} nor {\sc Gemini Ultra 1.0} 
solve any MCP problems correctly, while {\sc ChatGPT-4} solves only three. {\sc o1-preview} does better, solving 7 problems correctly. 
{\sc ChatGPT-4+Code} produces correct plans for 5 elaborations and occasionally generates solutions that are mostly correct but include minor issues that can be easily fixed manually. These cases are denoted with $\Delta$ in Table~\ref{tb:mcp_table}. 
{\sf LLM+AL}\footnote{{For {\sf LLM+AL} , we use {\sc o1-preview} (``o1-preview'' in the API) as the underlying LLM.}} automatically solves 7 MCP problems. For the remaining problems, an average of $3.1$ manual corrections to the generated $\bcp$ program are required to produce correct plans. 

We further include superficial variations of standard puzzles and observe that, even for these simple elaborations, the results are similar, as shown in Table~\ref{tb:cross_table}. These variations involve minor changes to the initial states (e.g., in Tower of Hanoi variations, disks are distributed among pegs in no particular order). The baseline LLMs perform poorly on these variations, with {\sc ChatGPT-4} and {\sc o1-preview} solving 1 and 2 variations correctly, respectively.  While {\sc ChatGPT-4+Code} performs better than the baseline LLMs, it still struggles with these variations. In contrast, {\sf LLM+AL} outperforms the baseline LLMs, and for the problems it fails to solve correctly, an average of only 2.2 corrections are required to produce the correct output. 

\smallskip
\noindent
\textbf{Effectiveness of Self-Revision.}\ \ \  
Self-Revision provides a notable improvement in the quality of the generated $\bcp$ programs. 
Across all 30 problems in Tables~\ref{tb:mcp_table} and \ref{tb:cross_table}, only 22.3\% (7/30) of programs generated prior to the Self-Revision step are executable without syntax errors. This percentage increases substantially to 86.7\% (26/30) after the Self-Revision step. 
Similarly, the proportion of programs that produce correct answers when run through the $\bcp$ reasoner rises from 16.6\% (5/30) before Self-Revision to 50\% (15/30) afterward.
In terms of issues requiring correction, there are 70 issues in the $\bcp$ programs prior to the Self-Revision step, but this number decreases to 42 following the Self-Revision step, with the detailed breakdown as explained in the next paragraph.

\begin{table}[ht] 
\small
    
    \centering
   \setlength\tabcolsep{3pt}
    \begin{tabular}{|c|c|c|c|c|c|c|c|c|c|}
        \hline
\multirow{2}{*}{Programs}        &  \multicolumn{9}{c|}{Issues} \\  \cline{2-10} 
        & \multicolumn{3}{c|}{Signature} & \multicolumn{3}{c|}{Rules} & \multicolumn{2}{c|}{Query} & Total \\ \cline{2-9}
     &  MD     & Syn      &    Sem       & MN      & HU      &  HN                        &  Syn      &  Sem  &   \\ \hline
Before &   18 & 11  &  3     & 16  & 5  & 10                     &  1  & 6 & 70 \\ \hline
After  &   7 & 1  &  4     & 12  & 5  & 8                     &  2  & 3 & 42 \\ \hline
    \end{tabular}
    \caption{Number of issues manually examined before (Before) and after Self-Revision (After).  The signature issues are missing sorts, object, variables, or constants in the declaration (MD), syntactic issues (Syn), or semantic issues (Sem). The rule issues are missing necessary rules (MN), harmful rules attempting to represent something not specified in the problem (HU), and harmful rules attempting to represent an aspect of the problem (HN). The query issues are either syntactic (Syn) or semantic (Sem).} \label{tb:corrections2}
\end{table}

\smallskip\noindent
\textbf{Program Issues.}\ \ \  
Table~\ref{tb:corrections2} enumerates all issues encountered in the final $\bcp$ programs produced by the {\sf LLM+AL} pipeline. Overall, we classify the 42 total issues into three categories: signature issues, rule issues, and query issues. Detailed descriptions of all issue cases are   provided in Appendix~D.

\smallskip
\noindent
\textit{Signature Issues.} There are 12 signature issues in total that can be categorized into three subcategories. The first and most common subcategory involves missing declarations for sorts, objects, variables, or constants, accounting for 7 cases. The second subcategory, comprising a single case, pertains to syntactically incorrect declarations. The third subcategory includes 4 cases and involve semantic issues in declarations, such as an incorrect supersort statement, which erroneously specify that certain objects are default members of another type.


\smallskip
\noindent
\textit{Rule Issues.} There are 25 rule issues in total that can be categorized into three subcategories. The first subcategory, comprising 48.0\% (12/25) of the issues, involves missing necessary rules required to solve the problems.
%
The second subcategory accounts for 20.0\% (5/25) of the rule issues and involves harmful rules that represent constraints or conditions not specified in the problem. For example, in MCP \#6 (where only one missionary and one cannibal can row), the pipeline generates an unnecessary rule that disallows both the missionary rower and the cannibal rower from being on the boat simultaneously.
The final subcategory accounts for 32.0\% (8/25) of rule issues and involves harmful rules that attempt to represent an aspect of the problem but do so incorrectly. For example, in MCP \#17 (where cannibals can become hungry), the pipeline generates an incorrect rule: ``On either bank, if there are missionaries present, the number of cannibals cannot exceed the number of missionaries." This rule fails to accurately reflect the elaboration, which specifies that the cannibals won’t become hungry as long as the strong missionary is rowing.
 



\smallskip
\noindent
\textit{Query Issues.} There are 5 query issues in total that can be categorized into two subcategories: Syntactic issues, which account for 2 cases, involve errors such as using invalid keywords like "initially" or "goal." Semantic issues, which account for 3 cases, pertain to incorrect initial and/or goal state conditions.

\subsection{Analysis} 


\smallskip
\noindent
\textbf{LLMs struggle to consistently adhere to state constraints.}\ \ \  
Of the 17 problems in Table~\ref{tb:mcp_table}, 14 are solvable, meaning they have valid plans to reach the goals. {\sc Claude 3 Opus} and {\sc Gemini 1.0 Ultra} fail to solve any of them correctly, while {\sc ChatGPT-4} produces 11 incorrect solutions. 
Among these incorrect plans, the state constraint—mandating that missionaries must not be outnumbered by cannibals—is frequently violated, despite clear instructions to adhere to it. Notably, 89.7\% (35 out of 39) of these violations occur within the first three steps of the plan.\footnote{All plans produced by baseline LLMs are available in the code repository.}
Even when these LLMs output intermediate states during plan generation, they fail to address apparent state constraint violations and continue generating flawed plans. In contrast, {\sc o1-preview} demonstrates a better adherence to constraints, producing fewer plans that violate state constraints and exhibiting more thoughtful consideration of the effects of actions and validation of states.
However, it occasionally refrains from generating plans for the problems known to be solvable, as discussed in the next paragraph. 

\smallskip\noindent
\textbf{LLMs do not reliably distinguish between solvable/unsolvable problems.}\ \ \ Some MCP elaborations are inherently unsolvable (MCP elaborations \#3, \#4, \#7). Even for these instances, 
{\sc ChatGPT-4}, {\sc Claude 3 Opus}, and {\sc Gemini 1.0 Ultra} generate (incorrect) plans, which can be considered hallucinations. Interestingly, these plans share some similarities. For example, in Elaborations \#3 and \#4, all plans include a state where cannibals outnumber missionaries. For Elaboration \#7, where no missionaries are allowed to row, all plans generated by these LLMs include a missionary rowing despite this being a clear violation of the elaboration. {\sc o1-preview} performs better in recognizing unsolvable problems, correctly identifying that no plans are possible for the three unsolvable elaborations. However, it is unclear how {\sc o1-preview} arrives at these conclusions, as its (partial) output does not seem to rule out all possibilities.\footnote{o1-preview does not show all of its CoT reasoning to users.} Moreover, it incorrectly concludes that no solution exists in five solvable instances (\#5, \#6, \#8, \#10, and \#17). Notably, {\sc o1-preview} is the only baseline model to explicitly claim that a solution is impossible, but it suffers from a high rate of false negatives. In contrast, {\sf LLM+AL} reliably ensures that there are no possible plans {\em up to certain fixed lengths} for Elaborations \#4 and \#7 automatically, and for Elaboration \#3 with minimal modifications, leveraging the formal semantics of $\bcp$ to validate its conclusions.\footnote{Strictly speaking, our pipeline is limited because it doesn't guarantee the non-existence of a plan of {\em arbitrary} length, for which one can use the method in Sec 6.6 of \cite{lee05a}.}

\smallskip\noindent
\textbf{LLM code generation struggles to adapt to problem elaborations.}\ \ \ 
We observe that {\sc {\sc ChatGPT-4+Code}} struggles to correctly incorporate new information into the Python programs it generates. In particular, it frequently fails to model actions accurately. As shown in Table~\ref{tb:mcp_table}, 5 of the 12 failures by {\sc ChatGPT-4+Code} on the elaborations (\#4, \#5, \#6, \#14, \#17) occur because the actions in the generated plans are underspecified. For example, some plans omit critical information, such as identifying who is rowing, which renders the plans unusable. 
The remaining 7 failures stem from a mix of issues: 2 precondition violations, 2 instances where no plans were generated, and 3 cases where unsolvable problems were not identified as such. This behavior is further corroborated by the results in Table~\ref{tb:cross_table}, where {\sc ChatGPT-4+Code} fails all problem variations except for one Sudoku example, despite correctly solving all the corresponding basic problems.


{\sc ChatGPT-4+Code} coincidentally succeeds in cases where it reuses the solution for the standard problem (MCP \#1, \#2, \#10, and \#11), producing identical plans for all. While the original solution happens to be valid for these elaborations, it may be suboptimal, as seen in \#10 and \#11. Notably, although these programs generate valid—and occasionally optimal—plans, they fail to accurately model the specific details introduced in the elaborations.
Additionally, {\sc ChatGPT-4+Code} exhibits an over-reliance on the standard problem's code, misapplying it to 5 MCP elaborations (\#5, \#6, \#7, \#13, and \#14). As in the coincidentally correct cases, {\sc ChatGPT-4+Code} ignores the relevant details specific to the elaborations and demonstrates a bias toward reproducing code for the basic MCP problem.
For MCP problems, all programs generated by CHATGPT-4+CODE use naive
 search algorithms: either breadth-first search (11 cases) or
 depth-first search (5 cases), often taking too long to find plans. In 9 instances, {\sc ChatGPT-4+Code} revises the code at least once.

\smallskip\noindent
\textbf{Declarative semantics of $\bcp$ works well with LLMs.}\ \ \ 
Unlike the LLM+Code approach, {\sf LLM+AL} does not require specifying which algorithms to use, thanks to the declarative semantics of action languages. The search algorithm implemented in the $\bcp$ reasoner is highly optimized for constraint satisfaction problems and can generate plans instantly.

\smallskip\noindent
\textbf{Self-Revision with the solver feedback significantly improves $\bcp$ program quality.}\ \ \  
Much like how a human might debug a program by testing its behavior and refining it based on feedback, Self-Revision enhances program quality through iterative refinement. It executes the program to verify satisfiability and employs simple queries generated by the LLM to ensure that the $\bcp$ program accurately models the problem. Additionally, it addresses both syntax and semantic errors by directly incorporating feedback from the $\bcp$ reasoner, progressively improving the program's correctness.
This approach significantly reduces the burden on the user, as many issues are resolved automatically. Even when some issues remain, they are relatively straightforward to address, further streamlining the process of creating accurate and robust $\bcp$ programs.


\smallskip\noindent
{\textbf{{\sf LLM+AL} benefits from human corrections, unlike LLM or LLM+Code.}} \ \ \
Occasionally requiring human corrections is a limitation of {\sc LLM+AL}, stemming from the fact that LLMs still make mistakes when generating action language representations. 
Despite this, these mistakes are relatively easy to correct due to the declarative semantics of $\bcp$. 
On the other hand, using only LLMs to solve these benchmarks fails to benefit from human corrections. We attempted 50 iterations of human corrections with {\sc ChatGPT-4} by indicating which parts of its answer were incorrect, but this did not help. For instance, in MCP elaboration \#13 (The Bridge), it repeatedly violated the constraint that cannibals should not outnumber missionaries.
Regarding {\sc ChatGPT-4+Code}, Python code is much less constrained and harder to interpret for action domains compared to $\bcp$, making it considerably more difficult for a human to correct the errors. 
\NB{Draw Table?}

\section{Conclusion} \label{sec:conclusion}

We propose {\sf LLM+AL}, a framework that bridges LLMs with action languages, enabling them to complement each other.  Compared to the direct use of LLMs, {\sf LLM+AL} achieves more robust and accurate reasoning about actions by leveraging the expressiveness and formal reasoning capabilities of action languages. While the generation of action language descriptions traditionally requires human expert knowledge, {\sf LLM+AL} simplifies this process through an automated process. Additionally, we employ a Self-Revision mechanism, an iterative approach in which an LLM generates sample queries to test the correctness of its previously generated $\bcp$ program. Based on feedback from the $\bcp$ reasoner, the LLM revises its program, significantly improving the quality of the final output.
While some mistakes may persist in the final programs, the generative capabilities of LLMs make creating action descriptions significantly easier compared to crafting them from scratch. It is likely that future LLM improvements will further reduce such errors. Moreover, fine-tuning LLMs could further enhance the performance of {\sf LLM+AL} provided it is feasible.



\section*{Acknowledgements}
We are grateful to the anonymous referees for their useful comments. 
This work was partially supported by the National Science Foundation under Grant IIS-2006747.



\appendix
\onecolumn 
{\bf \Large Appendix}

The appendix is organized as follows. 
Section~\ref{appendix:mcp_example} shows an example run for the basic Missionaries and Cannibals problem. Section~\ref{appendix:complete_prompts} includes all prompts used in {\sf LLM+AL}. Section~\ref{sec:baseline_prompts} shows the prompts used for the baseline LLM along with {\sc ChatGPT-4+Code}. Section~\ref{sec:error_cases} shows example error cases. Section~\ref{sec:mcp_elaborations} lists all the Missionaries and Cannibals elaborations that we use. 
All outputs by the baseline LLMs can be found in the code repository.

All experiments with {\sf LLM+AL} access model {\sf o1-preview} through OpenAI's API (``o1-preview''). All experiments are done on a machine with a I9-9900k processor, Nvidia Geforce RTX 2080 Ti, and 32 GB of RAM. Running {\sf LLM+AL} is mostly bottlenecked by calling the LLM, though the $\bcp$ reasoner occasionally requires significant time.

\section{The Pipeline Detailed with Missionaries and Cannibals Example} \label{appendix:mcp_example}

\subsection{Input}\label{appendix:bc_inputs}

\begin{tcolorbox}[fontupper=\small, title=Example Problem Input, label=input_ex]
Three missionaries and three cannibals shall cross a river in a boat which carries at most two people at once. The boat cannot cross the river with nothing on it. On either side, the cannibals may never outnumber the missionaries or the missionaries will be eaten. Also, the cannibals may never outnumber the missionaries on the boat or the missionaries will be eaten. \\

There are four types of things: "vessel", "location", "group", and "integer".

boat is a vessel.

bank1 and bank2 are locations.

missionaries and cannibals are groups.

0..10 are integers.\\

Actions

Crossing a vessel is an action.\\

Query

Three missionaries and three cannibals are at location bank1. The boat is at location bank1. Find a plan so that all missionaries and all cannibals are at location bank2.

\end{tcolorbox}

\subsection{${\cal BC}$+ Signature Generation}\label{appendix:bc_signature}
\newcommand{\Hilight}{\makebox[0pt][l]{\color{yellow}\rule[-4pt]{0.85\linewidth}{10pt}}}
\newcommand{\Hilighttwo}{\makebox[0pt][l]{\color{yellow}\rule[-4pt]{0.5\linewidth}{10pt}}}

\begin{lstlisting}[basicstyle=\scriptsize, escapechar=^]
Actions:
- cross(Vessel) - Cross the river in the Vessel.

Constants:
- loc(Vessel) - The location of the Vessel, remains the same unless explicitly changed, so it is an inertial fluent.
- numOnBank(Location, Group) - The number of people of Group at a Location (bank), remains the same unless explicitly changed, so it is an inertial fluent.
- numCrossing(Vessel, Group) - The number of people of Group crossing in the Vessel during the crossing action, an attribute of `cross(Vessel)`.

BC+ Signature:

:- sorts
    vessel;
    location;
    group;
    integer.

:- objects
    boat :: vessel;
    bank1, bank2 :: location;
    missionaries, cannibals :: group;
    0..10 :: integer.

:- variables
    V, V1, V2, V3, V4 :: vessel;
    L, L1, L2, L3, L4 :: location;
    G, G1, G2, G3, G4 :: group;
    N, N1, N2, N3, N4 :: integer;
    Bool, Bool1, Bool2, Bool3, Bool4 :: boolean.

:- constants
    cross(vessel) :: exogenousAction;
    numCrossing(vessel, group) :: attribute(integer) of cross(vessel);
    loc(vessel) :: inertialFluent(location);
    numOnBank(location, group) :: inertialFluent(integer).
\end{lstlisting}

\subsection{Knowledge Generation} \label{ssec:knowledge-generation-mcp}

\begin{lstlisting}
% Crossing the river in the vessel causes the location of the vessel to change to the opposite bank.
% Crossing the river in the vessel causes the number of people of each group on the departure bank to decrease by the number of people of that group crossing in the vessel.
% Crossing the river in the vessel causes the number of people of each group on the arrival bank to increase by the number of people of that group crossing in the vessel.
% The vessel cannot cross the river with zero people on it.
% The vessel cannot carry more than two people at once.
% The number of people of each group crossing in the vessel cannot exceed the number of people of that group on the bank where the vessel is located.
% The number of people of each group crossing in the vessel cannot be negative.
% After crossing, on either bank, the number of cannibals cannot exceed the number of missionaries unless the number of missionaries on that bank is zero.
% During crossing, the number of cannibals crossing in the vessel cannot exceed the number of missionaries crossing in the vessel unless the number of missionaries crossing is zero.
% The number of people of each group on either bank cannot be negative or exceed three.
\end{lstlisting}

\subsection{Rule and Query Generation}\label{appx:rule_generation}
At this point, ${\cal BC}$+ rules are generated based on the previously generated and revised knowledge:

\begin{lstlisting}
% constraints/rules

% 1. Crossing the river in the vessel causes the location of the vessel to change to the opposite bank.
cross(V) causes loc(V) = bank2 if loc(V) = bank1.
cross(V) causes loc(V) = bank1 if loc(V) = bank2.

% 2. Crossing the river in the vessel causes the number of people of each group on the departure bank to decrease by the number of people of that group crossing in the vessel.
cross(V) decrements numOnBank(L, G) by N if numCrossing(V, G) = N & loc(V) = L.

% 3. Crossing the river in the vessel causes the number of people of each group on the arrival bank to increase by the number of people of that group crossing in the vessel.
% 3.a If the vessel is at bank1, crossing increments numOnBank(bank2, G) by N.
cross(V) increments numOnBank(bank2, G) by N if numCrossing(V, G) = N & loc(V) = bank1.
% 3.b If the vessel is at bank2, crossing increments numOnBank(bank1, G) by N.
cross(V) increments numOnBank(bank1, G) by N if numCrossing(V, G) = N & loc(V) = bank2.

% 4. The vessel cannot cross the river with zero people on it.
nonexecutable cross(V) if N1 + N2 = 0 & N1 = numCrossing(V, missionaries) & N2 = numCrossing(V, cannibals).

% 5. The vessel cannot carry more than two people at once.
nonexecutable cross(V) if N1 + N2 > 2 & N1 = numCrossing(V, missionaries) & N2 = numCrossing(V, cannibals).

% 6. The number of people of each group crossing in the vessel cannot exceed the number of people of that group on the bank where the vessel is located.
nonexecutable cross(V) if N1 > N2 & N1 = numCrossing(V, G) & N2 = numOnBank(L, G) & loc(V) = L.

% 7. The number of people of each group crossing in the vessel cannot be negative.
nonexecutable cross(V) if N < 0 & N = numCrossing(V, G).

% 8. After crossing, on either bank, the number of cannibals cannot exceed the number of missionaries unless the number of missionaries on that bank is zero.
impossible numOnBank(L, cannibals) > numOnBank(L, missionaries) & numOnBank(L, missionaries) > 0.

% 9. During crossing, the number of cannibals crossing in the vessel cannot exceed the number of missionaries crossing in the vessel unless the number of missionaries crossing is zero.
nonexecutable cross(V) if N2 > N1 & N1 > 0 & N1 = numCrossing(V, missionaries) & N2 = numCrossing(V, cannibals).

% 10. The number of people of each group on either bank cannot be negative or exceed three.
impossible numOnBank(L, G) < 0.
impossible numOnBank(L, G) > 3.

% query
:- query
    0: numOnBank(bank1, missionaries) = 3 & numOnBank(bank1, cannibals) = 3 & numOnBank(bank2, missionaries) = 0 & numOnBank(bank2, cannibals) = 0 & loc(boat) = bank1;
    maxstep: numOnBank(bank2, missionaries) = 3 & numOnBank(bank2, cannibals) = 3.
\end{lstlisting}

\subsection{Initial ${\cal BC}$+ Program}\label{appendix:BC+ Program}
After the rule and query generation phase, an initial $\bcp$ program is generated as follows:
\begin{lstlisting}
:- sorts
    vessel;
    location;
    group;
    integer.


:- objects
    boat :: vessel;
    bank1, bank2 :: location;
    missionaries, cannibals :: group;
    0..10 :: integer.


:- variables
    V, V1, V2, V3, V4 :: vessel;
    L, L1, L2, L3, L4 :: location;
    G, G1, G2, G3, G4 :: group;
    N, N1, N2, N3, N4 :: integer;
    Bool, Bool1, Bool2, Bool3, Bool4 :: boolean.


:- constants
    cross(vessel) :: exogenousAction;
    numCrossing(vessel, group) :: attribute(integer) of cross(vessel);
    loc(vessel) :: inertialFluent(location);
    numOnBank(location, group) :: inertialFluent(integer).


% constraints/rules

% 1. Crossing the river in the vessel causes the location of the vessel to change to the opposite bank.
cross(V) causes loc(V) = bank2 if loc(V) = bank1.
cross(V) causes loc(V) = bank1 if loc(V) = bank2.

% 2. Crossing the river in the vessel causes the number of people of each group on the departure bank to decrease by the number of people of that group crossing in the vessel.
cross(V) decrements numOnBank(L, G) by N if numCrossing(V, G) = N & loc(V) = L.

% 3. Crossing the river in the vessel causes the number of people of each group on the arrival bank to increase by the number of people of that group crossing in the vessel.
% 3.a If the vessel is at bank1, crossing increments numOnBank(bank2, G) by N.
cross(V) increments numOnBank(bank2, G) by N if numCrossing(V, G) = N & loc(V) = bank1.
% 3.b If the vessel is at bank2, crossing increments numOnBank(bank1, G) by N.
cross(V) increments numOnBank(bank1, G) by N if numCrossing(V, G) = N & loc(V) = bank2.

% 4. The vessel cannot cross the river with zero people on it.
nonexecutable cross(V) if N1 + N2 = 0 & N1 = numCrossing(V, missionaries) & N2 = numCrossing(V, cannibals).

% 5. The vessel cannot carry more than two people at once.
nonexecutable cross(V) if N1 + N2 > 2 & N1 = numCrossing(V, missionaries) & N2 = numCrossing(V, cannibals).

% 6. The number of people of each group crossing in the vessel cannot exceed the number of people of that group on the bank where the vessel is located.
nonexecutable cross(V) if N1 > N2 & N1 = numCrossing(V, G) & N2 = numOnBank(L, G) & loc(V) = L.

% 7. The number of people of each group crossing in the vessel cannot be negative.
nonexecutable cross(V) if N < 0 & N = numCrossing(V, G).

% 8. After crossing, on either bank, the number of cannibals cannot exceed the number of missionaries unless the number of missionaries on that bank is zero.
impossible numOnBank(L, cannibals) > numOnBank(L, missionaries) & numOnBank(L, missionaries) > 0.

% 9. During crossing, the number of cannibals crossing in the vessel cannot exceed the number of missionaries crossing in the vessel unless the number of missionaries crossing is zero.
nonexecutable cross(V) if N2 > N1 & N1 > 0 & N1 = numCrossing(V, missionaries) & N2 = numCrossing(V, cannibals).

% 10. The number of people of each group on either bank cannot be negative or exceed three.
impossible numOnBank(L, G) < 0.
impossible numOnBank(L, G) > 3.


% query
:- query
    0: numOnBank(bank1, missionaries) = 3 & numOnBank(bank1, cannibals) = 3 & numOnBank(bank2, missionaries) = 0 & numOnBank(bank2, cannibals) = 0 & loc(boat) = bank1;
    maxstep: numOnBank(bank2, missionaries) = 3 & numOnBank(bank2, cannibals) = 3.
\end{lstlisting}

\subsection{Self-Revision: Final ${\cal BC}$+ Program}\label{appendix:BC+ Program revised}

\textbf{Satisfiability Check}
After running the initial program, the $\bcp$ reasoner does not compile the program due to a syntax error. The following feedback is provided:

\begin{lstlisting}
% Query 0: Check satisfiability of program signature and rules (ignores other queries). (satisfiable)
:- query = sat.

Cplus2ASP Output:

%ERROR: temp.bc:39:21]: The head of an additive law (G in "F decrements G") must be an additive constant (additiveFluent or additiveAction).

Since this fails, check the error message(s).
\end{lstlisting}

The LLM corrects this issue, re-writing the program as follows. The constant \texttt{numOnBank} is now correctly declared as an additive fluent.

\begin{lstlisting}
% BC+ signature
:- sorts
    vessel;
    location;
    group;
    integer.

:- objects
    boat :: vessel;
    bank1, bank2 :: location;
    missionaries, cannibals :: group;
    0..10 :: integer.

:- variables
    V, V1, V2, V3, V4 :: vessel;
    L, L1, L2, L3, L4 :: location;
    G, G1, G2, G3, G4 :: group;
    N, N1, N2, N3, N4 :: integer;
    Bool, Bool1, Bool2, Bool3, Bool4 :: boolean.

:- constants
    cross(vessel) :: exogenousAction;
    numCrossing(vessel, group) :: attribute(integer) of cross(vessel);
    loc(vessel) :: inertialFluent(location);
    numOnBank(location, group) :: additiveFluent(integer);

% Constraints/rules

% 1. Crossing the river in the vessel causes the location of the vessel to change to the opposite bank.
cross(V) causes loc(V) = bank2 if loc(V) = bank1.
cross(V) causes loc(V) = bank1 if loc(V) = bank2.

% 2. Crossing the river in the vessel causes the number of people of each group on the departure bank to decrease by the number of people of that group crossing in the vessel.
cross(V) decrements numOnBank(L, G) by N if numCrossing(V, G) = N & loc(V) = L.

% 3. Crossing the river in the vessel causes the number of people of each group on the arrival bank to increase by the number of people of that group crossing in the vessel.
% 3.a If the vessel is at bank1, crossing increments numOnBank(bank2, G) by N if numCrossing(V, G) = N & loc(V) = bank1.
cross(V) increments numOnBank(bank2, G) by N if numCrossing(V, G) = N & loc(V) = bank1.

% 3.b If the vessel is at bank2, crossing increments numOnBank(bank1, G) by N if numCrossing(V, G) = N & loc(V) = bank2.
cross(V) increments numOnBank(bank1, G) by N if numCrossing(V, G) = N & loc(V) = bank2.

% 4. The vessel cannot cross the river with zero people on it.
nonexecutable cross(V) if N1 + N2 = 0 & N1 = numCrossing(V, missionaries) & N2 = numCrossing(V, cannibals).

% 5. The vessel cannot carry more than two people at once.
nonexecutable cross(V) if N1 + N2 > 2 & N1 = numCrossing(V, missionaries) & N2 = numCrossing(V, cannibals).

% 6. The number of people of each group crossing in the vessel cannot exceed the number of people of that group on the bank where the vessel is located.
nonexecutable cross(V) if N1 > N2 & N1 = numCrossing(V, G) & N2 = numOnBank(L, G) & loc(V) = L.

% 7. The number of people of each group crossing in the vessel cannot be negative.
nonexecutable cross(V) if N < 0 & N = numCrossing(V, G).

% 8. After crossing, on either bank, the number of cannibals cannot exceed the number of missionaries unless the number of missionaries on that bank is zero.
impossible N1 > N2 & N2 > 0 & N1 = numOnBank(L, cannibals) & N2 = numOnBank(L, missionaries).

% 9. During crossing, the number of cannibals crossing in the vessel cannot exceed the number of missionaries crossing in the vessel unless the number of missionaries crossing is zero.
nonexecutable cross(V) if N2 > N1 & N1 > 0 & N1 = numCrossing(V, missionaries) & N2 = numCrossing(V, cannibals).

% 10. The number of people of each group on either bank cannot be negative or exceed three.
impossible N < 0 & N = numOnBank(L, G).
impossible N > 3 & N = numOnBank(L, G).

% Main query
:- query
    0: numOnBank(bank1, missionaries) = 3 & numOnBank(bank1, cannibals) = 3 & numOnBank(bank2, missionaries) = 0 & numOnBank(bank2, cannibals) = 0 & loc(boat) = bank1;
    maxstep: numOnBank(bank2, missionaries) = 3 & numOnBank(bank2, cannibals) = 3.

\end{lstlisting}

\noindent
\textbf{Sample Query Generation}
After the satisfiability check, the LLM generates the following sample queries.

\begin{lstlisting}
% Query 1: Have one missionary and one cannibal cross from bank1 to bank2 in the boat (satisfiable)
:- query
    0: numOnBank(bank1, missionaries) = 3 & numOnBank(bank1, cannibals) = 3 & loc(boat) = bank1 & numOnBank(bank2, missionaries) = 0 & numOnBank(bank2, cannibals) = 0;
    0: cross(boat) & numCrossing(boat, missionaries) = 1 & numCrossing(boat, cannibals) = 1.

% Query 2: Try to have two missionaries and one cannibal cross from bank1 to bank2 in the boat (unsatisfiable)
:- query
    0: numOnBank(bank1, missionaries) = 3 & numOnBank(bank1, cannibals) = 3 & loc(boat) = bank1 & numOnBank(bank2, missionaries) = 0 & numOnBank(bank2, cannibals) = 0;
    0: cross(boat) & numCrossing(boat, missionaries) = 2 & numCrossing(boat, cannibals) = 1.

% Query 3: Have one cannibal cross from bank1 to bank2 alone in the boat (satisfiable)
:- query
    0: numOnBank(bank1, missionaries) = 3 & numOnBank(bank1, cannibals) = 3 & loc(boat) = bank1 & numOnBank(bank2, missionaries) = 0 & numOnBank(bank2, cannibals) = 0;
    0: cross(boat) & numCrossing(boat, missionaries) = 0 & numCrossing(boat, cannibals) = 1.

% Query 4: Try to have one missionary cross from bank1 to bank2 alone in the boat (unsatisfiable)
:- query
    0: numOnBank(bank1, missionaries) = 3 & numOnBank(bank1, cannibals) = 3 & loc(boat) = bank1 & numOnBank(bank2, missionaries) = 0 & numOnBank(bank2, cannibals) = 0;
    0: cross(boat) & numCrossing(boat, missionaries) = 1 & numCrossing(boat, cannibals) = 0.

% Query 5: Try to have the boat cross from bank1 to bank2 with no people in it (unsatisfiable)
:- query
    0: numOnBank(bank1, missionaries) = 3 & numOnBank(bank1, cannibals) = 3 & loc(boat) = bank1 & numOnBank(bank2, missionaries) = 0 & numOnBank(bank2, cannibals) = 0;
    0: cross(boat) & numCrossing(boat, missionaries) = 0 & numCrossing(boat, cannibals) = 0.
\end{lstlisting}

\noindent
\textbf{Sample and Main Query Feedback}
And the feedback from running these queries, along with the main query is shown to the LLM as follows. 

\begin{lstlisting}
% Query 1: Have one missionary and one cannibal cross from bank1 to bank2 in the boat (satisfiable)
:- query
    0: numOnBank(bank1, missionaries) = 3 & numOnBank(bank1, cannibals) = 3 & loc(boat) = bank1 & numOnBank(bank2, missionaries) = 0 & numOnBank(bank2, cannibals) = 0;
    0: cross(boat) & numCrossing(boat, missionaries) = 1 & numCrossing(boat, cannibals) = 1.

Cplus2ASP Output:

Solving...
Solution: 1
	

	0:  loc(boat)=bank1 numOnBank(bank1, cannibals)=3 numOnBank(bank1, missionaries)=3 numOnBank(bank2, cannibals)=0 numOnBank(bank2, missionaries)=0

	ACTIONS:  cross(boat) numCrossing(boat, cannibals)=1 numCrossing(boat, missionaries)=1

	1:  loc(boat)=bank2 numOnBank(bank1, cannibals)=2 numOnBank(bank1, missionaries)=2 numOnBank(bank2, cannibals)=1 numOnBank(bank2, missionaries)=1

SATISFIABLE
Models       : 1


% Query 2: Try to have two missionaries and one cannibal cross from bank1 to bank2 in the boat (unsatisfiable)
:- query
    0: numOnBank(bank1, missionaries) = 3 & numOnBank(bank1, cannibals) = 3 & loc(boat) = bank1 & numOnBank(bank2, missionaries) = 0 & numOnBank(bank2, cannibals) = 0;
    0: cross(boat) & numCrossing(boat, missionaries) = 2 & numCrossing(boat, cannibals) = 1.

Cplus2ASP Output:

Solving...
UNSATISFIABLE
Models       : 0
No solution.



% Query 3: Have one cannibal cross from bank1 to bank2 alone in the boat (satisfiable)
:- query
    0: numOnBank(bank1, missionaries) = 3 & numOnBank(bank1, cannibals) = 3 & loc(boat) = bank1 & numOnBank(bank2, missionaries) = 0 & numOnBank(bank2, cannibals) = 0;
    0: cross(boat) & numCrossing(boat, missionaries) = 0 & numCrossing(boat, cannibals) = 1.

Cplus2ASP Output:

Solving...
Solution: 1
	

	0:  loc(boat)=bank1 numOnBank(bank1, cannibals)=3 numOnBank(bank1, missionaries)=3 numOnBank(bank2, cannibals)=0 numOnBank(bank2, missionaries)=0

	ACTIONS:  cross(boat) numCrossing(boat, cannibals)=1 numCrossing(boat, missionaries)=0

	1:  loc(boat)=bank2 numOnBank(bank1, cannibals)=2 numOnBank(bank1, missionaries)=3 numOnBank(bank2, cannibals)=1 numOnBank(bank2, missionaries)=0

SATISFIABLE
Models       : 1


% Query 4: Try to have one missionary cross from bank1 to bank2 alone in the boat (unsatisfiable)
:- query
    0: numOnBank(bank1, missionaries) = 3 & numOnBank(bank1, cannibals) = 3 & loc(boat) = bank1 & numOnBank(bank2, missionaries) = 0 & numOnBank(bank2, cannibals) = 0;
    0: cross(boat) & numCrossing(boat, missionaries) = 1 & numCrossing(boat, cannibals) = 0.

Cplus2ASP Output:

Solving...
UNSATISFIABLE
Models       : 0
No solution.



% Query 5: Try to have the boat cross from bank1 to bank2 with no people in it (unsatisfiable)
:- query
    0: numOnBank(bank1, missionaries) = 3 & numOnBank(bank1, cannibals) = 3 & loc(boat) = bank1 & numOnBank(bank2, missionaries) = 0 & numOnBank(bank2, cannibals) = 0;
    0: cross(boat) & numCrossing(boat, missionaries) = 0 & numCrossing(boat, cannibals) = 0.

Cplus2ASP Output:

Solving...
UNSATISFIABLE
Models       : 0
No solution.





:- query
    0: numOnBank(bank1, missionaries) = 3 & numOnBank(bank1, cannibals) = 3 & numOnBank(bank2, missionaries) = 0 & numOnBank(bank2, cannibals) = 0 & loc(boat) = bank1;
    maxstep: numOnBank(bank2, missionaries) = 3 & numOnBank(bank2, cannibals) = 3.


Cplus2ASP Output:

Solving...
Solution: 1
	

	0:  loc(boat)=bank1 numOnBank(bank1, cannibals)=3 numOnBank(bank1, missionaries)=3 numOnBank(bank2, cannibals)=0 numOnBank(bank2, missionaries)=0

	ACTIONS:  cross(boat) numCrossing(boat, cannibals)=2 numCrossing(boat, missionaries)=0

	1:  loc(boat)=bank2 numOnBank(bank1, cannibals)=1 numOnBank(bank1, missionaries)=3 numOnBank(bank2, cannibals)=2 numOnBank(bank2, missionaries)=0

	ACTIONS:  cross(boat) numCrossing(boat, cannibals)=1 numCrossing(boat, missionaries)=0

	2:  loc(boat)=bank1 numOnBank(bank1, cannibals)=2 numOnBank(bank1, missionaries)=3 numOnBank(bank2, cannibals)=1 numOnBank(bank2, missionaries)=0

	ACTIONS:  cross(boat) numCrossing(boat, cannibals)=2 numCrossing(boat, missionaries)=0

	3:  loc(boat)=bank2 numOnBank(bank1, cannibals)=0 numOnBank(bank1, missionaries)=3 numOnBank(bank2, cannibals)=3 numOnBank(bank2, missionaries)=0

	ACTIONS:  cross(boat) numCrossing(boat, cannibals)=1 numCrossing(boat, missionaries)=0

	4:  loc(boat)=bank1 numOnBank(bank1, cannibals)=1 numOnBank(bank1, missionaries)=3 numOnBank(bank2, cannibals)=2 numOnBank(bank2, missionaries)=0

	ACTIONS:  cross(boat) numCrossing(boat, cannibals)=0 numCrossing(boat, missionaries)=2

	5:  loc(boat)=bank2 numOnBank(bank1, cannibals)=1 numOnBank(bank1, missionaries)=1 numOnBank(bank2, cannibals)=2 numOnBank(bank2, missionaries)=2

	ACTIONS:  cross(boat) numCrossing(boat, cannibals)=1 numCrossing(boat, missionaries)=1

	6:  loc(boat)=bank1 numOnBank(bank1, cannibals)=2 numOnBank(bank1, missionaries)=2 numOnBank(bank2, cannibals)=1 numOnBank(bank2, missionaries)=1

	ACTIONS:  cross(boat) numCrossing(boat, cannibals)=0 numCrossing(boat, missionaries)=2

	7:  loc(boat)=bank2 numOnBank(bank1, cannibals)=2 numOnBank(bank1, missionaries)=0 numOnBank(bank2, cannibals)=1 numOnBank(bank2, missionaries)=3

	ACTIONS:  cross(boat) numCrossing(boat, cannibals)=1 numCrossing(boat, missionaries)=0

	8:  loc(boat)=bank1 numOnBank(bank1, cannibals)=3 numOnBank(bank1, missionaries)=0 numOnBank(bank2, cannibals)=0 numOnBank(bank2, missionaries)=3

	ACTIONS:  cross(boat) numCrossing(boat, cannibals)=2 numCrossing(boat, missionaries)=0

	9:  loc(boat)=bank2 numOnBank(bank1, cannibals)=1 numOnBank(bank1, missionaries)=0 numOnBank(bank2, cannibals)=2 numOnBank(bank2, missionaries)=3

	ACTIONS:  cross(boat) numCrossing(boat, cannibals)=0 numCrossing(boat, missionaries)=1

	10:  loc(boat)=bank1 numOnBank(bank1, cannibals)=1 numOnBank(bank1, missionaries)=1 numOnBank(bank2, cannibals)=2 numOnBank(bank2, missionaries)=2

	ACTIONS:  cross(boat) numCrossing(boat, cannibals)=1 numCrossing(boat, missionaries)=1

	11:  loc(boat)=bank2 numOnBank(bank1, cannibals)=0 numOnBank(bank1, missionaries)=0 numOnBank(bank2, cannibals)=3 numOnBank(bank2, missionaries)=3

SATISFIABLE
Models       : 1+


\end{lstlisting}

The LLM chooses not revise anything, as it marks ``[UNCHANGED]'' for the $\bcp$ signature and rules, main query, and sample queries.

\begin{lstlisting}
% PROGRAM CHANGED? [UNCHANGED]
% BC+ signature
:- sorts
    vessel;
    location;
    group;
    integer.


:- objects
    boat :: vessel;
    bank1, bank2 :: location;
    missionaries, cannibals :: group;
    0..10 :: integer.


:- variables
    V, V1, V2, V3, V4 :: vessel;
    L, L1, L2, L3, L4 :: location;
    G, G1, G2, G3, G4 :: group;
    N, N1, N2, N3, N4 :: integer;
    Bool, Bool1, Bool2, Bool3, Bool4 :: boolean.


:- constants
    cross(vessel) :: exogenousAction;
    numCrossing(vessel, group) :: attribute(integer) of cross(vessel);
    loc(vessel) :: inertialFluent(location);
    numOnBank(location, group) :: additiveFluent(integer).

% Constraints/rules

% 1. Crossing the river in the vessel causes the location of the vessel to change to the opposite bank.
cross(V) causes loc(V) = bank2 if loc(V) = bank1.
cross(V) causes loc(V) = bank1 if loc(V) = bank2.

% 2. Crossing the river in the vessel causes the number of people of each group on the departure bank to decrease by the number of people of that group crossing in the vessel.
cross(V) decrements numOnBank(L, G) by N if numCrossing(V, G) = N & loc(V) = L.

% 3. Crossing the river in the vessel causes the number of people of each group on the arrival bank to increase by the number of people of that group crossing in the vessel.
% 3.a If the vessel is at bank1, crossing increments numOnBank(bank2, G) by N if numCrossing(V, G) = N & loc(V) = bank1.
cross(V) increments numOnBank(bank2, G) by N if numCrossing(V, G) = N & loc(V) = bank1.

% 3.b If the vessel is at bank2, crossing increments numOnBank(bank1, G) by N if numCrossing(V, G) = N & loc(V) = bank2.
cross(V) increments numOnBank(bank1, G) by N if numCrossing(V, G) = N & loc(V) = bank2.

% 4. The vessel cannot cross the river with zero people on it.
nonexecutable cross(V) if N1 + N2 = 0 & N1 = numCrossing(V, missionaries) & N2 = numCrossing(V, cannibals).

% 5. The vessel cannot carry more than two people at once.
nonexecutable cross(V) if N1 + N2 > 2 & N1 = numCrossing(V, missionaries) & N2 = numCrossing(V, cannibals).

% 6. The number of people of each group crossing in the vessel cannot exceed the number of people of that group on the bank where the vessel is located.
nonexecutable cross(V) if N1 > N2 & N1 = numCrossing(V, G) & N2 = numOnBank(L, G) & loc(V) = L.

% 7. The number of people of each group crossing in the vessel cannot be negative.
nonexecutable cross(V) if N < 0 & N = numCrossing(V, G).

% 8. After crossing, on either bank, the number of cannibals cannot exceed the number of missionaries unless the number of missionaries on that bank is zero.
impossible N1 > N2 & N2 > 0 & N1 = numOnBank(L, cannibals) & N2 = numOnBank(L, missionaries).

% 9. During crossing, the number of cannibals crossing in the vessel cannot exceed the number of missionaries crossing in the vessel unless the number of missionaries crossing is zero.
nonexecutable cross(V) if N2 > N1 & N1 > 0 & N1 = numCrossing(V, missionaries) & N2 = numCrossing(V, cannibals).

% 10. The number of people of each group on either bank cannot be negative or exceed three.
impossible N < 0 & N = numOnBank(L, G).
impossible N > 3 & N = numOnBank(L, G).


% MAIN QUERY CHANGED? [UNCHANGED]
% Three missionaries and three cannibals are at location bank1. The boat is at location bank1. Find a plan so that all missionaries and all cannibals are at location bank2. 
:- query
    0: numOnBank(bank1, missionaries) = 3 & numOnBank(bank1, cannibals) = 3 & numOnBank(bank2, missionaries) = 0 & numOnBank(bank2, cannibals) = 0 & loc(boat) = bank1;
    maxstep: numOnBank(bank2, missionaries) = 3 & numOnBank(bank2, cannibals) = 3.


% SAMPLE QUERIES CHANGED? [UNCHANGED]
% Query 1: Have one missionary and one cannibal cross from bank1 to bank2 in the boat (satisfiable)
:- query
    0: numOnBank(bank1, missionaries) = 3 & numOnBank(bank1, cannibals) = 3 & loc(boat) = bank1 & numOnBank(bank2, missionaries) = 0 & numOnBank(bank2, cannibals) = 0;
    0: cross(boat) & numCrossing(boat, missionaries) = 1 & numCrossing(boat, cannibals) = 1.

% Query 2: Try to have two missionaries and one cannibal cross from bank1 to bank2 in the boat (unsatisfiable)
:- query
    0: numOnBank(bank1, missionaries) = 3 & numOnBank(bank1, cannibals) = 3 & loc(boat) = bank1 & numOnBank(bank2, missionaries) = 0 & numOnBank(bank2, cannibals) = 0;
    0: cross(boat) & numCrossing(boat, missionaries) = 2 & numCrossing(boat, cannibals) = 1.

% Query 3: Have one cannibal cross from bank1 to bank2 alone in the boat (satisfiable)
:- query
    0: numOnBank(bank1, missionaries) = 3 & numOnBank(bank1, cannibals) = 3 & loc(boat) = bank1 & numOnBank(bank2, missionaries) = 0 & numOnBank(bank2, cannibals) = 0;
    0: cross(boat) & numCrossing(boat, missionaries) = 0 & numCrossing(boat, cannibals) = 1.

% Query 4: Try to have one missionary cross from bank1 to bank2 alone in the boat (unsatisfiable)
:- query
    0: numOnBank(bank1, missionaries) = 3 & numOnBank(bank1, cannibals) = 3 & loc(boat) = bank1 & numOnBank(bank2, missionaries) = 0 & numOnBank(bank2, cannibals) = 0;
    0: cross(boat) & numCrossing(boat, missionaries) = 1 & numCrossing(boat, cannibals) = 0.

% Query 5: Try to have the boat cross from bank1 to bank2 with no people in it (unsatisfiable)
:- query
    0: numOnBank(bank1, missionaries) = 3 & numOnBank(bank1, cannibals) = 3 & loc(boat) = bank1 & numOnBank(bank2, missionaries) = 0 & numOnBank(bank2, cannibals) = 0;
    0: cross(boat) & numCrossing(boat, missionaries) = 0 & numCrossing(boat, cannibals) = 0.
\end{lstlisting}

The final correct program is generated as follows.

\begin{lstlisting}
:- sorts
    vessel;
    location;
    group;
    integer.



:- objects
    boat :: vessel;
    bank1, bank2 :: location;
    missionaries, cannibals :: group;
    0..10 :: integer.



:- variables
    V, V1, V2, V3, V4 :: vessel;
    L, L1, L2, L3, L4 :: location;
    G, G1, G2, G3, G4 :: group;
    N, N1, N2, N3, N4 :: integer;
    Bool, Bool1, Bool2, Bool3, Bool4 :: boolean.



:- constants
    cross(vessel) :: exogenousAction;
    numCrossing(vessel, group) :: attribute(integer) of cross(vessel);
    loc(vessel) :: inertialFluent(location);
    numOnBank(location, group) :: additiveFluent(integer).

% Constraints/rules

% 1. Crossing the river in the vessel causes the location of the vessel to change to the opposite bank.
cross(V) causes loc(V) = bank2 if loc(V) = bank1.
cross(V) causes loc(V) = bank1 if loc(V) = bank2.

% 2. Crossing the river in the vessel causes the number of people of each group on the departure bank to decrease by the number of people of that group crossing in the vessel.
cross(V) decrements numOnBank(L, G) by N if numCrossing(V, G) = N & loc(V) = L.

% 3. Crossing the river in the vessel causes the number of people of each group on the arrival bank to increase by the number of people of that group crossing in the vessel.
% 3.a If the vessel is at bank1, crossing increments numOnBank(bank2, G) by N if numCrossing(V, G) = N & loc(V) = bank1.
cross(V) increments numOnBank(bank2, G) by N if numCrossing(V, G) = N & loc(V) = bank1.

% 3.b If the vessel is at bank2, crossing increments numOnBank(bank1, G) by N if numCrossing(V, G) = N & loc(V) = bank2.
cross(V) increments numOnBank(bank1, G) by N if numCrossing(V, G) = N & loc(V) = bank2.

% 4. The vessel cannot cross the river with zero people on it.
nonexecutable cross(V) if N1 + N2 = 0 & N1 = numCrossing(V, missionaries) & N2 = numCrossing(V, cannibals).

% 5. The vessel cannot carry more than two people at once.
nonexecutable cross(V) if N1 + N2 > 2 & N1 = numCrossing(V, missionaries) & N2 = numCrossing(V, cannibals).

% 6. The number of people of each group crossing in the vessel cannot exceed the number of people of that group on the bank where the vessel is located.
nonexecutable cross(V) if N1 > N2 & N1 = numCrossing(V, G) & N2 = numOnBank(L, G) & loc(V) = L.

% 7. The number of people of each group crossing in the vessel cannot be negative.
nonexecutable cross(V) if N < 0 & N = numCrossing(V, G).

% 8. After crossing, on either bank, the number of cannibals cannot exceed the number of missionaries unless the number of missionaries on that bank is zero.
impossible N1 > N2 & N2 > 0 & N1 = numOnBank(L, cannibals) & N2 = numOnBank(L, missionaries).

% 9. During crossing, the number of cannibals crossing in the vessel cannot exceed the number of missionaries crossing in the vessel unless the number of missionaries crossing is zero.
nonexecutable cross(V) if N2 > N1 & N1 > 0 & N1 = numCrossing(V, missionaries) & N2 = numCrossing(V, cannibals).

% 10. The number of people of each group on either bank cannot be negative or exceed three.
impossible N < 0 & N = numOnBank(L, G).
impossible N > 3 & N = numOnBank(L, G).



% Three missionaries and three cannibals are at location bank1. The boat is at location bank1. Find a plan so that all missionaries and all cannibals are at location bank2. 
:- query
    0: numOnBank(bank1, missionaries) = 3 & numOnBank(bank1, cannibals) = 3 & numOnBank(bank2, missionaries) = 0 & numOnBank(bank2, cannibals) = 0 & loc(boat) = bank1;
    maxstep: numOnBank(bank2, missionaries) = 3 & numOnBank(bank2, cannibals) = 3.


\end{lstlisting}

A correct plan is generated as output.

\begin{lstlisting}
Solving...
Solution: 1
	

0:  loc(boat)=bank1 numOnBank(bank1, cannibals)=3 numOnBank(bank1, missionaries)=3 numOnBank(bank2, cannibals)=0 numOnBank(bank2, missionaries)=0

ACTIONS:  cross(boat) numCrossing(boat, cannibals)=2 numCrossing(boat, missionaries)=0

1:  loc(boat)=bank2 numOnBank(bank1, cannibals)=1 numOnBank(bank1, missionaries)=3 numOnBank(bank2, cannibals)=2 numOnBank(bank2, missionaries)=0

ACTIONS:  cross(boat) numCrossing(boat, cannibals)=1 numCrossing(boat, missionaries)=0

2:  loc(boat)=bank1 numOnBank(bank1, cannibals)=2 numOnBank(bank1, missionaries)=3 numOnBank(bank2, cannibals)=1 numOnBank(bank2, missionaries)=0

ACTIONS:  cross(boat) numCrossing(boat, cannibals)=2 numCrossing(boat, missionaries)=0

3:  loc(boat)=bank2 numOnBank(bank1, cannibals)=0 numOnBank(bank1, missionaries)=3 numOnBank(bank2, cannibals)=3 numOnBank(bank2, missionaries)=0

ACTIONS:  cross(boat) numCrossing(boat, cannibals)=1 numCrossing(boat, missionaries)=0

4:  loc(boat)=bank1 numOnBank(bank1, cannibals)=1 numOnBank(bank1, missionaries)=3 numOnBank(bank2, cannibals)=2 numOnBank(bank2, missionaries)=0

ACTIONS:  cross(boat) numCrossing(boat, cannibals)=0 numCrossing(boat, missionaries)=2

5:  loc(boat)=bank2 numOnBank(bank1, cannibals)=1 numOnBank(bank1, missionaries)=1 numOnBank(bank2, cannibals)=2 numOnBank(bank2, missionaries)=2

ACTIONS:  cross(boat) numCrossing(boat, cannibals)=1 numCrossing(boat, missionaries)=1

6:  loc(boat)=bank1 numOnBank(bank1, cannibals)=2 numOnBank(bank1, missionaries)=2 numOnBank(bank2, cannibals)=1 numOnBank(bank2, missionaries)=1

ACTIONS:  cross(boat) numCrossing(boat, cannibals)=0 numCrossing(boat, missionaries)=2

7:  loc(boat)=bank2 numOnBank(bank1, cannibals)=2 numOnBank(bank1, missionaries)=0 numOnBank(bank2, cannibals)=1 numOnBank(bank2, missionaries)=3

ACTIONS:  cross(boat) numCrossing(boat, cannibals)=1 numCrossing(boat, missionaries)=0

8:  loc(boat)=bank1 numOnBank(bank1, cannibals)=3 numOnBank(bank1, missionaries)=0 numOnBank(bank2, cannibals)=0 numOnBank(bank2, missionaries)=3

ACTIONS:  cross(boat) numCrossing(boat, cannibals)=2 numCrossing(boat, missionaries)=0

9:  loc(boat)=bank2 numOnBank(bank1, cannibals)=1 numOnBank(bank1, missionaries)=0 numOnBank(bank2, cannibals)=2 numOnBank(bank2, missionaries)=3

ACTIONS:  cross(boat) numCrossing(boat, cannibals)=0 numCrossing(boat, missionaries)=1

10:  loc(boat)=bank1 numOnBank(bank1, cannibals)=1 numOnBank(bank1, missionaries)=1 numOnBank(bank2, cannibals)=2 numOnBank(bank2, missionaries)=2

ACTIONS:  cross(boat) numCrossing(boat, cannibals)=1 numCrossing(boat, missionaries)=1

11:  loc(boat)=bank2 numOnBank(bank1, cannibals)=0 numOnBank(bank1, missionaries)=0 numOnBank(bank2, cannibals)=3 numOnBank(bank2, missionaries)=3

SATISFIABLE
Models       : 1+
Calls        : 1
Time         : 1.129s (Solving: 0.18s 1st Model: 0.18s Unsat: 0.00s)
CPU Time     : 1.129s


\end{lstlisting}

\section{Complete Prompts} \label{appendix:complete_prompts}

Capital words encapsulated like ``$<$WORDS$>$'' represent problem specific text, such as problem inputs, or outputs from previously generated steps. Lowercase words encapsulated like ``$<$words$>$'' are not problem specific and are used to reduce verbosity. The prompt templates and exact inputs to the LLM are found in the code repository.

A $\bcp$ description is put in all prompts, except the prompt for knowledge generation, since the knowledge generated is in natural language instead of any formal representation.

\begin{lstlisting}
Action language BC+ consists of (1) a signature, which is a finite set of propositional atoms, and (2) causal laws. An atom is true, false, or of the form c=v, where c is a constant, and v is an element in its domain. Constants c are divided into two groups: fluent constants and action constants. A formula is a propositional combination of atoms. A fluent formula is a formula such that all constants occurring in it are fluent constants. An action formula is a formula that contains at least one action constant and no fluent constants. 

For example, a fluent constant "loc" which represents the location of some object, may have a value from {l1, l2, l3}. To express the location is l2, one would write:
loc = l2.

Similarly, an action constant "drive" representing driving a car, may have a value from {true, false} (though in general actions can be non-boolean). To represent this, one could write:
drive = true.

For boolean constants c, c is shorthand for c=true, and ~c is shorthand for c=false. Thus "drive = true" can be represented simply as "drive". The standard logical operators are: "-" for negation, "&" for conjunction, "|" for disjunction, and "->" for implication.

Numeric values may combine with unary operators "-" (negative), "abs" (absolute value, e.g., abs(n)), and binary operators "+" (addition), "-" (subtraction), "*" (multiplication), "//" (floor division), and "mod" (modulus, e.g., n mod m). Comparison is done with "=" (equality), "\=" (inequality), "<" (less than), ">" (greater than), "<=" (less than or equal), and ">=" (greater than or equal).

An action description D is a set of causal laws that define a transition system, which can be represented as a directed graph. In this graph, the vertices correspond to states of the world, while the edges signify transitions between these states.

[CAUSAL LAWS: BASIC FORM]

In BC+, there are three types of basic laws: (1) static law, (2) action dynamic law, and (3) fluent dynamic law:

STATIC LAW:

	F if G,

where F and G are fluent formulas. This expresses that F is true if the formula G is true.

ACTION DYNAMIC LAW:

	F if G,

where F is an action formula and G is a formula. This expresses that action formula F is true if G is true.

FLUENT DYNAMIC LAW:

	F if G after H,

where F and G are fluent formulas, and H is a formula. This expresses that F is true if G is true after H is true.

[CAUSAL LAWS: SOME SHORTHAND ABBREVIATIONS]

BC+ allows several kinds of shorthand notation for the basic causal laws. These make it convenient to directly express commonsense knowledge.


CAUSES:

	a causes F if H,

where a is an action, F is a fluent formula, H is a formula, is shorthand for the fluent dynamic law:

	F after a & H.

This rule intuitively expresses that an action a causes some effect F if condition H is true.

IMPOSSIBLE:

	impossible F,

where F is a fluent formula, is shorthand for the static law:

	false if F.

This rule intuitively expresses that formula F should be false in every state.

NONEXECUTABLE:

	nonexecutable a_1 & ... & a_k if G,

where a_1 & ... & a_k are action constants and G is a formula, is shorthand for the fluent dynamic law:

	false if true after a_1 & ... & a_k & G.

This rule intuitively expresses that actions a_1 & ... & a_k all being true is not possible if G is true.

DEFAULT:

	default c=v if F,

where F is a fluent formula, stands for a static causal law. 

This intuitively expresses that by default c=v holds when F is true in the state. If there is evidence to the contrary, then c=v may not hold. 

	default c=v if F after G,

where F is a fluent formula, and G is a formula, and stands for a fluent dynamic law which intuitively expresses that by default c=v holds when F is true in the state and F was true in the previous state. If there is evidence to the contrary, then c=v may not hold.

ALWAYS:

	always F,

where F is a formula which can contain both fluent and action constants, is shorthand for the fluent dynamic law:

	false if true after ~F.

This rule intuitively expresses that every transition must satisfy F.

[SORT/OBJECT/VARIABLE/CONSTANT DECLARATION IMPLEMENTATION]

Action language BC+ is implemented in a program as follows. A BC+ program consists of a sort declaration, object declaration, variable declaration, constant declaration, along with a set of causal laws.

SORT DECLARATION:
A sort is a named set of elements which is used to specify the domain of each constant and variable. First the sort is declared
using a sort declaration statement and, later, is defined by adding objects to it in an object declaration statement.

Example:
:- sorts
    int;
    box.

This declares the sorts: int, and box.

A sort can also be a supersort, that is, a sort s can automatically include the objects within another sort s2, denoted s >> s2.

:- sorts
    package >> box.
This declares the sort package and box, where package is a supersort and includes the sort box.

OBJECT DECLARATION:
An object is a value in a sort which a constant can take. It is also used
in parameter lists to construct nested objects and sets of constants. 

Example:
:- objects
    1..3 :: int;
    o(int, int) :: box.

This declares 1,2, and 3 as objects within int, and the objects:
o(1, 1), o(1, 2), o(1, 3),
o(2, 1), o(2, 2), o(2, 3),
o(3, 1), o(3, 2), and o(3, 3),
as values within box.

Consider the following sort and object declaration:

:- sorts
    package >> box.

:- objects
    p1, p2 :: package;
    b1, b2, b3 :: box.

Since package is a supersort of box, there are 5 objects that are of sort package: p1, p2, b1, b2, and b3. There are only 3 objects of sort box: b1, b2 and b3.

VARIABLE DECLARATION:
A variable is a placeholder symbol which will be replaced with each object in its domain during grounding.

Example:
:- variables
    I, I1, I2 :: int;
    B, B1, B2 :: box.
This declares the variables I, I1, and I2 over the objects within int, and variables B, B1, and B2 range over the objects within box.
Variables are denoted as capital letters which are replaced with all possible values in the domain of their associated sort in a process called grounding.

CONSTANT DECLARATION:
Similar to object symbols, constants are defined within a constant declaration statement and have a base identifier, an optional list of parameter sort, and a sort which makes up the constant's domain. In addition, they have a constant declaration type, such as "exogenousAction" or "inertialFluent". Inertial fluents are common, and used for constants whose value persists through time unless affected otherwise.

Example:
:- constants
    p(int) :: inertialFluent(boolean);
    move(box) :: exogenousAction.
This declares the inertial fluent p which takes an int argument and is a boolean value, and the action move, which takes a box argument.

[CAUSAL LAWS BASIC FORM IMPLEMENTATION]

STATIC LAW EXAMPLE:

	F if G,

where F and G are fluent formulas, is written for example:

% The location of a person is the same as the car if the person is in the car.
loc(person)=L if inCar & loc(car)=L.

ACTION DYNAMIC LAW EXAMPLE:

	F if G,

where F is an action formula and G is a formula, is written for example:

% If A is assigned to location Loc, and it is ready, then moveTo(A) is true.
moveTo(A) if assigned(A) = Loc & ready(A),
where moveTo(A) is an action, and assigned(A) and ready(A) are fluent constants. Here moveTo is a boolean action constant.

FLUENT DYNAMIC LAW EXAMPLE:

	F if G after H,

where F and G are fluent formulas, and H is a formula, is written for example:

% If you enter the car after the car is unlocked, then you are in the car.
in(car) if unlocked(car) after enter(car).

[CAUSAL LAW SHORTHAND ABBREVIATIONS IMPLEMENTATION]

CAUSES EXAMPLE:

An expression of the form
	a causes F if H,

where a is an action, F is a fluent formula, H is a formula, is written for example:
% Driving the car to a location causes the car to be at that location of the car has gas.
drive(car, loc) causes location(car) = loc if hasGas(car)

IMPOSSIBLE EXAMPLE:

An expression of the form

	impossible F,

where F is a fluent formula, is written for example:

% The cat and mouse cannot be at the same location
impossible loc(cat)=Loc1 & loc(mouse) = Loc2 & Loc1=Loc2.

Recall that F in "impossible F" must be a fluent formula, meaning that F cannot contain action constants, including attributes.

The following two examples are incorrect, where drive(car) is an action, and numPassengers(car) and numDrivers(car) are attributes:
% The car cannot be driven if it has no fuel.
impossible drive(car) & fuel(car) = 0.

% The combined number of passengers and drivers cannot exceed 5.
impossible N1 + N2 > 5 & N1 = numPassengers(car) & N2 = numDrivers(car).

These are wrong because drive(car) is an action constant, and the attributes numPassengers and numDrivers are action constants, and they are in an impossible constraint.

NONEXECUTABLE EXAMPLE:

An expression of the form:

	nonexecutable a1 & a2 & ... & an if G, 

where a_1 & ... & a_k are action constants and G is a formula, is written for example:

% It is not permissible to drive a bus and a car if they are driving to the same location. 
nonexecutable driveCar(Loc1) & driveBus(Loc2) if Loc1=Loc2.

% It is not permissible to lift an object if it is heavy.
nonexecutable lift(object) if heavy(object).

On the other hand, the following:

nonexecutable lifted(object) if heavy(object),

is not allowed when lifted(object) is a fluent and not an action.

DEFAULT EXAMPLES:

An expression of the form

	default c=v if F,

where F is a fluent formula, is written for example as:

% By default, the power is on if the switch is turned.
default powerOn if switchTurned.

An expression of the form

	default c=v if F after G,

where F is a fluent formula, and G is a formula, is written for example as:

% The ferry will be on the right bank if it has gas after it was on the left bank, and no evidence suggests otherwise.
default ferry(right) if hasGas after ferry(left)

ALWAYS EXAMPLE:

	always F,

where F is a formula which can contain both fluent and action constants, is written for example:

% the number departing from a location is always less than or equal to the number at the location
always departing(Loc) <= numAt(Loc).

where departing/1 is an action constant and numAt/1 is a fluent.
Note: There should not be an "if" in a "always" rule, instead use "&".

Keep in mind that "always F" should be used for conditions which should be true for every transition.

[EXTENDED CONSTANT DECLARATION]
We can expand constant declaration to allow for attributes, additive fluents and additive actions.

ATTRIBUTES:
Attributes are types of action constants which are tied to actions, and given additional information about an action. For instance:

:- constants
    move(block) :: exogenousAction;
    destination(block) :: attribute(location) of move(block).

This declaration tells us that destination(block) is an attribute of the action of moving a block, which has value of location, representing the block's destination.

Important: Attributes should have as their first arguments, the same arguments (and in the same order) as the action which they are an attribute of.
For example, any attribute of the action "fly(airplane,location)" should have the form "attributeName(airplane,location,...)", after which the action arguments "airplane" and "location", can other arguments be listed.
An attribute which represents the number of passengers on board of a nationality can be written as:
onBoard(airplane,location,nationality) :: attribute(integer) of fly(airplane,location).

Where the following would be incorrect:
Incorrect: passengers(nationality) :: attribute(integer) of fly(airplane,location),
since passengers should instead have the arguments "airplane" and "location" before "nationality".

Important: If an attribute is ever declared, then the action it is an attribute of must have at least one argument.

For example, the above declaration would be incorrect if it was the following:

:- constants
    move :: exogenousAction;
    destination :: attribute(location) of move.

Here, since destination is an attribute of move, but move has no argument, this is wrong.

ADDITIVE FLUENTS AND ADDITIVE ACTIONS:

There are special constants additive fluents and additive actions, which are useful for modeling the effects of concurrent actions.
An additive fluent is a fluent (tied to states) with numerical values such that the effect of several concurrently executed actions on it can be computed by adding the effects of the individual actions.
An additive action constant (tied to a transition) is similar except that it can handle numeric values of concurrent actions during transitions.

With a buy and sell action, and numExchanged attribute of exchange, additive fluents and additive actions can be declared as follows:
:- constants
    buy(item) :: exogenousAction;
    sell(item) :: exogenousAction;
    exchange(item) :: exogenousAction;
    numExchanged(item) :: attribute(integer) of exchange(item); 
    inStock(item) :: additiveFluent(integer);
    netChange(item) :: additiveAction(integer).

inStock(item) represents the the number of an item in stock, which can be affected by both buy(item) and sell(item). netChange(item) represents the net number of an item which is exchanged or bought/sold during a given transition or event.

[EXTENDED CAUSAL LAW SHORTHAND ABBREVIATIONS: INCREMENT/DECREMENT LAWS IMPLEMENTATION]

INCREMENT/DECREMENT LAWS:
Additive fluents can only be updated in increment/decrement laws, which are expressed as:

    a increments c by v if G,

where a is an action, c is an additive fluent, v is an integer, and G is a fluent, is used to express that some additive fluent (c), should increase by some value v if G is true and action a happens.

Similarly:

    a decrements c by v if G,

is used to decrease the additive constant c by value v if G is true and action a happens.

Examples:
% buy(coin) increments has(coin) by N if howMany(coin,N).
% buy(coin) decrements has(money) by N*M if howMany(coin,N) & price(coin)=M.

Additive actions similarly are only updated in increment/decrement statements:

    a increments c by v if G.
    a decrements c by v if G.

where c is an additive action.
While additive fluents are used to describe state information, additive actions are about events/transitions. For example, suppose multiple actions can occur at once for a domain involving buying and selling, and the additive action netExchanged is declared:
% buy(coin) increments netExchanged(coin) by N if howMany(coin,N).
% sell(coin) decrements netExchanged(coin) by N if howMany(coin,N).

This is useful for tracking numerical values during a transition, which may be used, e.g., if there should never be a negative amount of net coins exchanged at a given time, then netExchanged(coin) can be used to construct such a law.

In general, if something causes something to increase/decrease, then we use increment/decrement. Updating the value of an additive fluent or action can only be done in an increment or decrement law (e.g., "a causes c=2" is not allowed if "c" represents an additive fluent/action).

[ARITHMETIC WITH ATTRIBUTES AND ADDITIVE CONSTANTS]
Attributes and additive actions should not occur in the operands of arithmetics such as addition, subtraction, multiplication, floor division, or modulus operators. They must be set to variables, and the variables must be used as the operands.

The following two rules are incorrect, where weight(truck,...) is an attribute:
% A truck cannot be driven if the combined weight of the cargo (C), passengers (P), and fuel (F), is greater than the truck's capacity.
nonexecutable drive(truck, P, C) if weight(truck, C) + weight(truck, P) + weight(truck, F) > capacity(truck).

% Driving the truck causes the weight of the truck's fuel to decrease by 15.
drive(truck, P, C) causes weight(truck, fuel) = weight(truck, fuel) - 15.

These are incorrect because the attributes weight(truck,...) are directly used in arithmetic.
Instead, these rules should be written as follows:
% A truck cannot be driven if the combined weight of the cargo (C), passengers (P), and fuel (F), is greater than the truck's capacity.
nonexecutable drive(truck, P, C) if N1 + N2 + N3 > capacity(truck) &  weight(truck, C) = N1 & weight(truck, P) = N2 & weight(truck, F) = N3.

% Driving the truck causes the weight of the truck's fuel to decrease by 15.
drive(truck, P, C) causes weight(truck, fuel) = N1 - 15 if N1 = weight(truck, fuel) & N2.

[RESTRICTION OF NESTED CONSTANTS]
A constant cannot appear as an argument of another constant. Instead, one can introduce a variable in place of a nested constant.

For example, the following is wrong since the (boolean) constant roadBlock(...), contains the nested constant loc(C):
nonexecutable move(C) if roadBlock(loc(C)).

Instead, loc(C) should be set to a variable, like the following:
nonexecutable move(C) if roadBlock(L) & L = loc(C).

[VARIABLE QUANTIFICATION]
When using "always F", make sure to quantify over variables if needed.
For example, say we want to express that the combined weight of passengers and the truck is less than the capacity of the truck. The following would be incorrect:
always weight(passengers) = N1 & weight(truck) = N2 & capacity(truck) = N3 & N1 + N2 < N3.

This is because of a quantification issue. This rule informally reads as "For all values N1, N2, and N3, the weight of the passengers is N1, the weight of the truck is N2, the capacity of the truck is N3, and N1 + N2 < N3." Since this will be grounded for every possible value that N1 can take it, there will be contradiction. Furthermore, the condition "N1 + N2 < N3" will obviously not hold over all possible values of the variables N1, N2, and N3.

To correctly write this, first set the variables in the antecedent of an implication:

always (weight(passengers) = N1 & weight(truck) = N2 & capacity(truck) = N3) -> N1 + N2 < N3.

This informally reads as "If the weight of the passengers is N1, the weight of the truck is N2, and the capacity of the truck is N3, then N1 + N2 < N3", and aligns with what we want to express.

\end{lstlisting}

\begin{figure*}[ht]
\begin{center}
\includegraphics[width=.9\columnwidth]{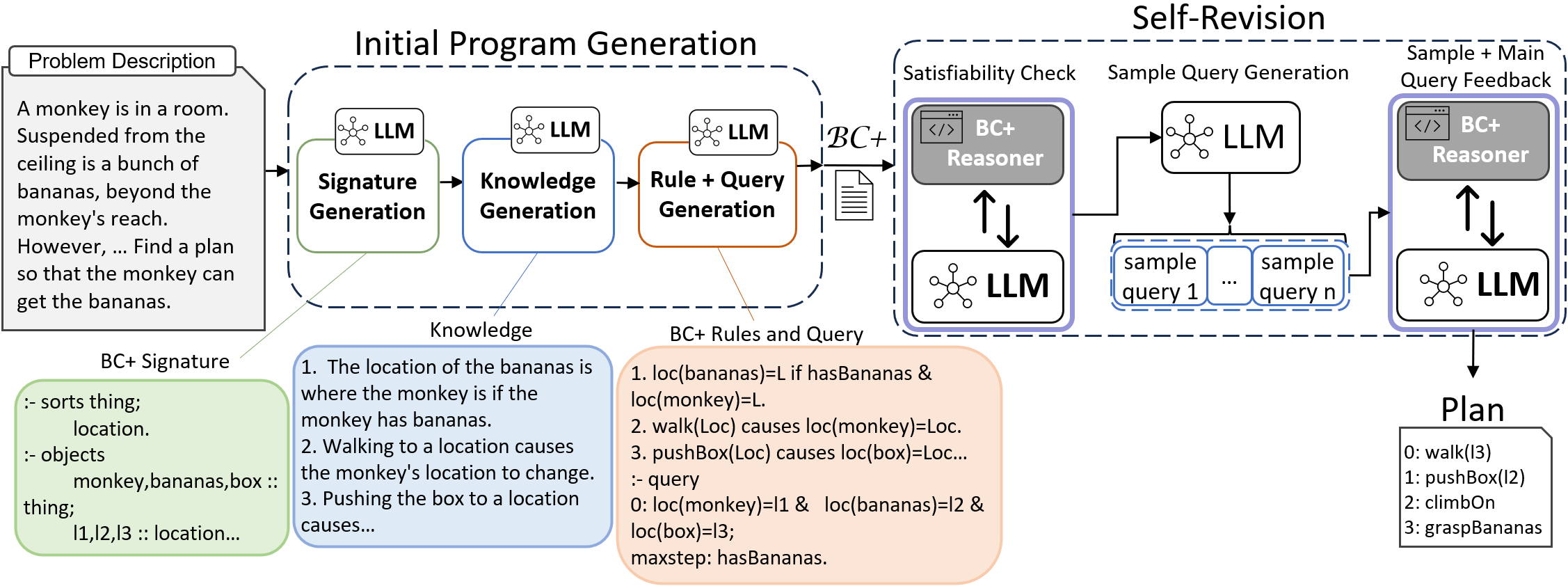}
\end{center}
\caption{{\sf LLM+AL}}
\label{fig:main_fig_appx}
\end{figure*}

\subsection{$\bcp$ Signature Generation}\label{appendix:signature_gen}
This prompt is used in  the Signature Generation portion of Initial Program Generation, as shown in Figure~\ref{fig:main_fig_appx}.


\begin{lstlisting}
<conceptual description of bc+>


[INSTRUCTIONS]
Given a hint, problem description, and knowledge about a domain, produce the relevant fluent and action constants and the BC+ signature. Do not use the same names for constants, even if they are of a different type or have different arguments. For example, on(block) and on(block,table) cannot both be declared.

<few-shot examples>

[Problem]
<DOMAIN>


As shown in the examples, write the actions, constants, and BC+ signature. Make sure to include enough variables for each sort, which will be needed later to write rules (5 variables each should be enough, for example, N, N1, N2, N3, N4 for integer, etc.).
Encapsulate your answer in 3 backticks like the following:

```
Actions:
<ACTIONS>

Constants:
<CONSTANTS AND THEIR READING>

BC+ Signature:
<SIGNATURE>
```
\end{lstlisting}

\subsection{Knowledge Generation}\label{appendix:knowledge_gen}
This prompt is used in the Knowledge Generation portion of Initial Program Generation, as shown in Figure~\ref{fig:main_fig_appx}.

\begin{lstlisting}
Given a description of a problem, write knowledge about the domain that we would expect to represent in an action language BC+ program. These should represent logical constraints and cause and effect in the problem.

For example, a task about moving objects would involve some knowledge about what is required to move, what the effect of moving an object does, such as the location of the object changing or its previous location now being empty. These restrictions and effects are dependent on the context of the problem. Here we list some example problems and extract knowledge.

<few-shot examples>

[Problem]
Description:
<PROBLEM DESCRIPTION> Provide the relevant knowledge. Only use the terminology in the hint/description to write the knowledge.

Hint:
<signature description>

BC+ description
<ACTIONS AND CONSTANTS>

<BC+ Signature>

Generate the the knowledge relevant to this problem, ONLY using natural langauge. Do not mention any of the program within this knowledge, though the signature should be considered when writing the knowledge. Encapsulate it in three backticks like so:

```
% generated knowledge
% <knowledge #1>
% <knowledge #2>
...

\end{lstlisting}

\subsection{Rule Generation}\label{appendix:rule_gen}
This prompt is used in the Rule and Query Generation portion of Initial Program Generation, as shown in Figure~\ref{fig:main_fig_appx}.

\begin{lstlisting}
<conceptual description of bc+>

QUERIES:
Queries have three components: an optional label, an optional maximum step which list formulas which should be true in the last step of the plan, and a set of constraint formulas to apply, each parametrized with the step at which they should be applied.
For example the following query represents an initial state and an action at step 0, and a set of constraints which should be met at the end of the plan.

:- query
	label :: test_query;
	0: loc(b1)=b2 & loc(b2)=b3 & loc(b3)=a & loc(b4)=b & loc(b5)=c;
	0: move(b1, b5);
	maxstep: loc(b5,c) & loc(b4,b5) & loc(b3,b4) & loc(b2,b3) & loc(b1,b2).

<few-show examples>

[INSTRUCTIONS]

Based on the problem description, previously generated BC+ signature and natural language rules/constraints to represent in BC+, generate the rules/constraints in BC+ along with the query.

[PROBLEM]

Description:
<PROBLEM DESCRIPTION>

Hint:
<HINT>

BC+ description:
<ACTIONS AND CONSTANTS>

<DOMAIN>

Rules/constraints to represent in BC+:
<CONSTRAINTS>

Query (in natural language):
<QUERY>

Do not alter or change the BC+ signature in any way, only present the constraints/rules (numbered), and query together in the following format, encapsulated in three backticks:

```
% constraints/rules
% 1. <comment for rule 1>
<BC+ rule 1>

% 2. <comment for rule 2>
<BC+ rule 1>
...

% query
< BC+ query>

```
\end{lstlisting}

\subsection{Satisfiability Check} \label{ssec:satisfiability check}
This prompt is used by the LLM in the Satisfiability Check portion of Self-Revision, as shown in Figure~\ref{fig:main_fig_appx}.

\begin{lstlisting}
<conceptual description of bc+>

[INSTRUCTIONS]
Consider the problem and BC+ program, which when run with a basic query to check satisfiability (independent of the main query), fails.

[Problem]
<PROBLEM DESCRIPTION>

BC+ Program:
<BC+ PROGRAM>

Main query:
<QUERY>
<BC+ QUERY>

Feedback:
<FEEDBACK>

Based on the feedback, revise the program (signature and/or rules), and possibly the main query if needed to match a new signature. The format should be like the following, with everything together, encapsulated in 3 backticks:

```% BC+ signature
<Enter BC+ signature>

% Generated constraints
<Enter BC+ rules/constraints>

% Main query
<Enter the main query>```
\end{lstlisting}

\subsection{Sample Query Generation} \label{ssec:sample query generation}
This prompt is used by the LLM in the Sample Query Generation portion of Self-Revision, as shown in Figure~\ref{fig:main_fig_appx}.

\begin{lstlisting}
<conceptual description of bc+>

QUERIES:
Queries have three components: an optional label, an optional maximum step which list formulas which should be true in the last step of the plan, and a set of constraint formulas to apply, each parametrized with the step at which they should be applied.
For example the following query represents an initial state and an action at step 0, and a set of constraints which should be met at the end of the plan.

:- query
	label :: test_query;
	0: loc(b1)=b2 & loc(b2)=b3 & loc(b3)=a & loc(b4)=b & loc(b5)=c;
	0: move(b1, b5);
	maxstep: loc(b5,c) & loc(b4,b5) & loc(b3,b4) & loc(b2,b3) & loc(b1,b2).

[INSTRUCTIONS]

Based on the problem description, and generated BC+ program. Generate some simple queries which can be used to make sure the program is working properly. For example, consider the following example.

<few-shot examples>

[PROBLEM]

Description:
<PROBLEM DESCRIPTION>

BC+ description:
<ACTIONS AND CONSTANTS>

<DOMAIN>

Rules/constraints to represent in BC+:
<CONSTRAINTS>

Main Query:
<QUERY>
<BC+ QUERY>

Based on the problem description, and generated BC+ program. Generate some simple queries which can be used to make sure the program is working properly. Only generate a few, no more than 5. Append either (satisfiable) or (unsatisfiable) to the end of the query in natural language, depending on whether the query should work or not. Keep in mind, the starting states of the sample queries should be allowable based on the already written constraints in the program (e.g., if a query comes back unsatisfiable, it shouldn't be because the start state was unsatisfiable). Do so in the following format, encapsulated in 3 backticks:
```
% Query 1: <natural language query> (satisfiable/unsatisfiable)
<BC+ Query>

% Query 2: <natural language query> (satisfiable/unsatisfiable)
<BC+ Query>
...

```
\end{lstlisting}

\subsection{Sample and Main Query Feedback} \label{ssec:sample and main query feedback}
This prompt is used by the LLM in the Sample and Main Query Feedback portion of Self-Revision, as shown in Figure~\ref{fig:main_fig_appx}.

\begin{lstlisting}
<conceptual description of bc+>

QUERIES:
Queries have three components: an optional label, an optional maximum step which list formulas which should be true in the last step of the plan, and a set of constraint formulas to apply, each parametrized with the step at which they should be applied.
For example the following query represents an initial state and an action at step 0, and a set of constraints which should be met at the end of the plan.

:- query
	label :: test_query;
	0: loc(b1)=b2 & loc(b2)=b3 & loc(b3)=a & loc(b4)=b & loc(b5)=c;
	0: move(b1, b5);
	maxstep: loc(b5,c) & loc(b4,b5) & loc(b3,b4) & loc(b2,b3) & loc(b1,b2).

<few-shot examples>

[CHECKLIST]
The following is a checklist which should be considered when writing and revising a BC+ program.
1. In general, if some conjunction of fluents cannot be true, then we use "impossible" when writing the rules, but if instead we want to assert some constant is caused to be false, then we use the negation (~). For example, if it is impossible for an object to be on the table and under it, we might write "impossible onTable(object) & underTable(object).", but express that no rain causes the ground to not be wet, we can write "~groundWet if noRain".
2. "impossible F" can only be used with fluents and "&". Do not use attributes or action constants in laws of the form "impossible F". "nonexecutable" is about actions not being permissible if certain fluents hold and/or certain actions are performed.
3. Additive constants can only be updated in increment/decrement laws.
4. In "a causes G if H", "G", recall G should be a fluent formula. Therefore, it should not directly contain action constants (including attributes). If one wants to set some constant to the value of an attribute, then the fluent formula G should contain a variable represnting the attribute, rather than the attribute itself.

For example, if the attribute nextDestination(Loc) is not set to a variable first, then this will be a malformed law error, since there is an action constant in G:
drive(car) causes loc(car) = nextDestination(Loc) if loc(boat) = Loc.

Instead, this should be written as:
drive(car) causes loc(car) = Loc2 if Loc2 = nextDestination(Loc) & loc(boat) = Loc.

This works because nextDestination(Loc) is first set to a variable.
5. Only use variables which are declared, if new variables are introduced, they must be declared.
6. In shorthand laws "default c=v if F" and "default c=v if F after G", "c" cannot be an additive constant, since this is can update the value of them, but they should only be updated in increment/decrement laws.
7. "always" shorthand abbreviations should not contain "if", but only conjunctions ("&").
8. In the case that a disjunction is used in the body of a rule, encapsulate it in parentheses, e.g., (loc(city1) | loc(city2)).
9. Do not nest constants in arguments of other constants. For example, do not use the following: nonexecutable move(C) if roadBlock(loc(C)). Instead, loc(C) should be set to a variable: nonexecutable move(C) if roadBlock(L) & L = loc(C).
10. Make sure that the query is exhaustive. For example, if the initial condition is that bus stop A has 10 people, and the goal condition is that the bus stop B has 10 people, it may need to be specified that bus stop B has 0 people in the initial condition, or the BC+ solver can just set the initial state to be the goal state.
11. Make sure to only use constants which are declared in the constant declaration. If new constans are used, then they must be added in the declaration.
12. Make sure when writing rules, that variables representing constants are of the correct sort, i.e., the sort of the variable is the same as the sort of the value of the constant it is representing.
For example, if there is a constant "capacity(vehicle)" which is of type "inertialFluent(integer)", then only variables of sort "integer" should be used to represent the constant in rules.
13. Don't use simple, or rigid fluents in the constant declaration. Use "inertialFluent" instead.
14. Don't use non-specified fluents. For example, there should not be constant declarations like "country(location) :: country", since this will be recognized as a rigid fluent.
15. Keep in mind that "always F" should be used for conditions which should be true for every transition and shout not be used to express conditions for a state. 

For example, consider a domain about trading coins that has the action constant numExchanged(coin), which is an attribute of some trading action, and the fluent constant amount(coin). numExchanged(coin) represents the net number of coins exchanged in a transition, from possibly multiple trade actions. amount(coin) represents the number of coins in a state (before and after trades).

If one wants to express that the net number of exchanged coins during an action is smaller than some limit for every trade, this can be expressed with:
always netExchanged(coin) < limit(coin).

However, if one wants to express that the amount of coins does not exceed some limit, this cannot be expressed with "always F". The following is incorrect:
always amount(coin) <= limit(coin).

This wrong because if a state does not have a transition to another state, then the condition "amount(coin) <= limit(coin)" can be false. For example, in a transition system with a single state, or the last state in a transition system.

Instead, it can be expressed as:
impossible amount(coint) > limit(coin).

This enforces the following is true for every state.
16. Attributes and additive actions should not occur in the operands of arithmetics such as addition, subtraction, multiplication, floor division, or modulus operators. They must be set to variables, and the variables must be used as the operands.

[INSTRUCTIONS]
Consider the problem and BC+ program, and simple sample queries which were ran to check the basic functionality of the program. The output of each sample query and the main query is presented. Check that the Cplus2ASP output makes sense.

[Problem]
<PROBLEM DESCRIPTION>

BC+ Program:
<BC+ PROGRAM>

Sample queries and outputs:
<FEEDBACK>

Main query and output:
% Original natural language query: <QUERY>
<FEEDBACK MAIN QUERY>

[INSTRUCTIONS (continued)]
1) [FEEDBACK OUTLINE]
Based on the feedback, if any of the following segments are incorrect, then mark them with "[CHANGED]", and revise them:
- PROGRAM
- MAIN QUERY
- SAMPLE QUERIES

Otherwise, if a segment is correct, mark it with "[UNCHANGED]", and simply copy their contents. 

2) [SAMPLE QUERY GUIDANCE]
Keep in mind, the outputs from running the sample queries is to help check that the program is working properly. In some cases, the sample queries themselves may be wrong (e.g., a syntax error), and can be re-written. The sample queries marked (satisfiable) should be correctly be satisfiable and the queries marked (unsatisfiable) should be unsatisfiable. There should not be apparent violations in the constraints, and the action preconditions/effects should enforced properly. If there are, then revise the main program. Overall, changes should be succint, e.g., only change relevant parts while leaving the remainder alone.

3) [MAIN QUERY OUTPUT]
IMPORTANT: Check that each action in the final plan of the main query makes sense given the state before it, and that the resulting state correctly follows.
If something doesn't align with the common sense relative to the problem description, then update the program. Though less likely, if you suspect there is an issue with the main BC+ query, it may be changed, but it should have the original meaning of the main query as shown under "Main query and output", after "Original natural language query: ". The query itself should not contain actions, but have only the initial state and the final state (indicated with "maxstep").

4) [FORMAT]
The format should be like the following, encapsulated in 3 backticks:

```% PROGRAM CHANGED? [CHANGED/UNCHANGED]
% BC+ signature
<Enter BC+ signature>

% Generated constraints
<Enter BC+ rules/constraints>

% MAIN QUERY CHANGED? [CHANGED/UNCHANGED]
<Enter the main query>

% SAMPLE QUERIES CHANGED? [CHANGED/UNCHANGED]
<Enter the sample queries (do NOT include the Cplus2ASP outputs)>
```
\end{lstlisting}

\section{Baseline Prompts}\label{sec:baseline_prompts}

\subsection{Baseline LLM ({\sc ChatGPT-4}, {\sc Claude 3 Opus}, {\sc Gemini 1.0 Ultra}, {\sc o1-preview})}
The format of the prompt used is:
\begin{lstlisting}
<PROBLEM DESCRIPTION>

<QUERY>
\end{lstlisting}
For example, for the River Cross domain:
\begin{lstlisting}
There are four types of things: vessel, location, group, and integer.

Problem Description:
Three missionaries and three cannibals shall cross a river in a boat which carries at most two people at once. The boat cannot cross the river with nothing on it. On either side, the cannibals may never outnumber the missionaries or the missionaries will be eaten. Also, the cannibals may never outnumber the missionaries on the boat or the missionaries will be eaten. The boat's capacity is 2.

boat is a vessel. bank1 and bank2 are locations. missionaries and cannibals are groups. 0, 1, 2, 3, 4, 5, 6, 7, 8, 9, and 10 are integers.

Three missionaries and three cannibals are at location bank1. The boat is at location bank1. Find a plan so that all missionaries and all cannibals are at location bank2.
\end{lstlisting}

\subsection{LLM + Code ({\sc ChatGPT-4+Code})}
\begin{lstlisting}
Use the code interpreter to generate a Python program to solve the following problem. Revise the program as many times as needed until the program is correct.

<PROBLEM DESCRIPTION>

<QUERY>
\end{lstlisting}

\section{LLM + AL Error Cases} \label{sec:error_cases}
The following constains all issues from {\sf LLM+AL} after Self-Revision. Unless otherwise stated, all issues shown contribute 1 issue to the total number of issues.
\subsection{Signature}

\textbf{Missing sort, object, variable, or constant in the declaration.}

Sudoku1 - the sort ``int'' needs objects instantiated from 0 to 8 (written ``0..8''). This is required because arithmetic is done with variables of the sort ``int'', so there must exist objects (representing integers) of ``int''.
\begin{lstlisting}[basicstyle=\scriptsize, escapechar=^]
:- objects
    0..9 :: value0;
    1..9 :: value;
    0..8 :: row;
    0..8 :: column.
    ^{\cred \% Missing: 0..8 :: int. }^
\end{lstlisting}

Sudoku (var2) - There are missing object declarations for the row, column and integer.
\begin{lstlisting}[basicstyle=\scriptsize, escapechar=^]
^{\cred :- objects \% Missing}^
    ^{\cred 1..9 :: row; \% Missing}^
    ^{\cred 1..9 :: column;  \% Missing}^
    ^{\cred 0..9 :: integer. \% Missing}^
\end{lstlisting}

Sudoku (var2) - There is a missing sort declaration for \texttt{integer}.
\begin{lstlisting}[basicstyle=\scriptsize, escapechar=^]
:- sorts
    row;
    column;
    value;
    ^{\cred integer. \% Missing}^
\end{lstlisting}

MCP \#6 (not everyone can row) - There is a missing constant which tracks the location of rowers, \texttt{rowersAt(basicGroup)}.
\begin{lstlisting}[basicstyle=\scriptsize, escapechar=^]
:- constants
    cross(vessel) :: exogenousAction;
    numCrossing(vessel, basicGroup) :: attribute(integer) of cross(vessel);
    rower(vessel, basicGroup) :: attribute(boolean) of cross(vessel);
    ^{\cred rowerAt(basicGroup) :: inertialFluent(location); \% Missing}^
    capacity(vessel) :: inertialFluent(integer);
    boatLoc :: inertialFluent(location);
    numAt(basicGroup, place) :: additiveFluent(integer).
\end{lstlisting}

MCP \#9 (big cannibal and small missionary) - Variables \texttt{L1, L2, N1,} and \texttt{N2} were used in rules, but never declared. Thus they are missing from the signature.
\begin{lstlisting}[basicstyle=\scriptsize, escapechar=^]
:- variables
    M, M1, M2, M3 :: missionary;
    C, C1, C2, C3 :: cannibal;
    P, P1, P2 :: person;
    Loc, Loc1, Loc2, NewLoc :: place;
    N_m, N_c, N :: integer;
    ^{\cred L1, L2, N1, N2 :: integer. \% Missing }^
\end{lstlisting}

MCP \#14 (the boat leaks with two people on it) - The action \texttt{bail} is missing from the constant declaration.
\begin{lstlisting}[basicstyle=\scriptsize, escapechar=^]
:- constants
    cross(vessel) :: exogenousAction;
    from(vessel) :: attribute(location) of cross(vessel);
    to(vessel) :: attribute(location) of cross(vessel);
    numOnBoard(vessel, group) :: attribute(integer) of cross(vessel);
    loc(vessel) :: inertialFluent(location);
    numOnBank(group, location) :: inertialFluent(integer);
    numOnBoat(group) :: inertialFluent(integer);
    capacity(vessel) :: inertialFluent(integer);
    ^{\cred bail(vessel) :: exogenousAction. \% Missing }^
\end{lstlisting}

MCP \#17 (cannibals can become hungry) - A fluent constant which represents whether the cannibals are hungry is missing.
\begin{lstlisting}[basicstyle=\scriptsize, escapechar=^]
:- constants
    loc(vessel) :: inertialFluent(location);  % The location of the vessel.
    capacity(vessel) :: inertialFluent(integer);  % The capacity of the vessel.
    cross(vessel) :: exogenousAction;  % Crossing action.
    numCrossing(vessel, group) :: attribute(integer) of cross(vessel);  % Number from a group crossing in the vessel.
    toLocation(vessel) :: attribute(location) of cross(vessel);  % Destination location.
    numAt(group, location) :: additiveFluent(integer);  % Number of people of a group at a location.
    ^{\cred     hungry :: inertialFluent(boolean). \% Missing }^
\end{lstlisting}

\noindent
\textbf{Syntax Issues.}
MCP \# 19 (there are two sets of groups) - Variables are incorrectly used in the constant declaration. This is syntactically wrong, only sorts should be used as arguments. \texttt{S, G,} and \texttt{L} should be changed to \texttt{set, group,} and \texttt{location}.

\begin{lstlisting}[basicstyle=\scriptsize, escapechar=^]
:- constant
cross(^{\cred S}^) :: exogenousAction;
dest(^{\cred S}^) :: attribute(location) of cross(^{\cred S}^);
numCrossing(^{\cred S}^, ^{\cred G}^) :: attribute(integer) of cross(^{\cred S}^);
boatLoc(^{\cred S}^) :: inertialFluent(location);
numAt(^{\cred S}^, ^{\cred G}^, ^{\cred L}^) :: additiveFluent(integer).
\end{lstlisting}

\noindent
\textbf{Semantic Issues.}
MCP \# 8 (a very big cannibal must cross alone) - The following sort declaration incorrectly states that missionaries and cannibals are supersorts of each other. Rather, there should be semicolons after the first occurrence of \texttt{missionary} and \texttt{cannibal}.

\begin{lstlisting}[basicstyle=\scriptsize, escapechar=^]
:- sorts
integer;
^{\cred missionary,}^
^{\cred cannibal,}^
person >> missionary, cannibal;
vessel;
location.
\end{lstlisting}

MCP \#9 (big cannibal and small missionary) - (1) The supersort statement is incorrect. \texttt{person} should be a supersort of missionary and cannibal, not the opposite. (2) \texttt{vessel} should not be a \texttt{place}, thus the supersort statement \texttt{place >> location, vessel} is wrong.
\begin{lstlisting}[basicstyle=\scriptsize, escapechar=^]
:- sorts
^{\cred missionary $>>$ person; \% should be person $>>$ missionary}^
^{\cred  cannibal $>>$ person; \% should be person $>>$ cannibal }^
vessel;
location;
^{\cred  place $>>$ location, vessel; \% should be just place $>>$ location }^
thing >> person, vessel;
integer.
\end{lstlisting}

MCP \#5 (an oar on each bank) - The constants \texttt{numMissionariesOn}, \texttt{numCannibalsOn}, and \texttt{numOarsOn}, should be attributes rather than inertial fluents.
\begin{lstlisting}[basicstyle=\scriptsize, escapechar=^]
:- constants
    cross(vessel) :: exogenousAction;
    loc(vessel) :: inertialFluent(location);
    capacity(vessel) :: inertialFluent(integer);
    numMissionariesAt(location) :: inertialFluent(integer);
    numCannibalsAt(location) :: inertialFluent(integer);
    numOarsAt(location) :: inertialFluent(integer);
    numMissionariesOn(vessel) :: ^{\cred inertialFluent(integer); \% should be an attribute}^
    numCannibalsOn(vessel) :: ^{\cred inertialFluent(integer); \% should be an attribute}^
    numOarsOn(vessel) :: ^{\cred inertialFluent(integer). \% should be an attribute}^
\end{lstlisting}

\subsection{Rule}

\textbf{Missing necessary rules.}
Tower of Hanoi (3, var1) - The constraint which says that two disks cannot be at the same location is missing:

\begin{lstlisting}[basicstyle=\scriptsize, escapechar=^]
impossible loc(D) = L & loc(D1) = L & D\=D1.
\end{lstlisting}

Sudoku (var2) - The following state constraints are required. This is counted as 3 separate issues.
\begin{lstlisting}[basicstyle=\scriptsize, escapechar=^]
% No two cells in the same row have the same value
false if val(R, C) = V & val(R, C1) = V & C \= C1.

% No two cells in the same column have the same value
false if val(R, C) = V & val(R1, C) = V & R \= R1.

% No two cells in the same 3x3 block have the same value
false if val(R, C) = V & val(R1, C1) = V & (R \= R1 | C \= C1) &
(R // 3 = R1 // 3) & (C // 3 = C1 // 3).
\end{lstlisting}

MCP \#6 (only one missionary and one cannibal can row) - There is a missing rule required to update the rower locations.
\begin{lstlisting}[basicstyle=\scriptsize, escapechar=^]
cross(boat) causes rowerAt(BG) = L if boatLoc = L1 & rower(boat, BG) & L\=L1.
\end{lstlisting}

MCP \#6 (only one missionary and one cannibal can row) - There is a missing rule required which disallows the rower to cross if it is not where the boat is.
\begin{lstlisting}[basicstyle=\scriptsize, escapechar=^]
nonexecutable cross(boat) if rower(boat, BG) & rowerAt(BG) \= L & boatLoc = L.
\end{lstlisting}

MCP \#8 (a very big cannibal must cross alone) - The are missing rules to represent that the cannibals cannot outnubmer the missionaries.
\begin{lstlisting}[basicstyle=\scriptsize, escapechar=^]
% disallow 3 cannibals and 2 missionaries
impossible loc(M) = Loc & loc(M1) = Loc & loc(M2) = Loc1 & M\=M1 &
loc(C) = Loc & loc(C1) = Loc & loc(C2) = Loc & C\=C1 & C\=C2 & C1\=C2 &
Loc \= Loc1.

% disallow 3 cannibals and 1 missionary
impossible loc(M) = Loc & loc(M1) = Loc1 & loc(M2) = Loc1 & M1\=M2 &
loc(C) = Loc & loc(C1) = Loc & loc(C2) = Loc & C\=C1 & C\=C2 & C1\=C2 &
Loc \= Loc1.

% disallow 2 cannibals and 1 missionary
impossible loc(M) = Loc & loc(M1) = Loc1 & loc(M2) = Loc1 & M1\=M2 &
loc(C) = Loc & loc(C1) = Loc & loc(C2) = Loc1 & C\=C1 & C\=C2 & C1\=C2 &
Loc \= Loc1.
\end{lstlisting}

MCP \# 14 (the boat leaks with two people on it) - The constraint which disallows crossing to your current location is missing.
\begin{lstlisting}[basicstyle=\scriptsize, escapechar=^]
nonexecutable cross(V) if to(V) = loc(V).
\end{lstlisting}

MCP \# 14 (the boat leaks with two people on it) - The rule which ensures bailing is done when two people are on board is missing is missing.
\begin{lstlisting}[basicstyle=\scriptsize, escapechar=^]
bail(V) if numOnBoard(V, missionaries) = N1 & numOnBoard(V, cannibals) = N2 & N1 + N2 = 2.
\end{lstlisting}

MCP \#17 (cannibals can become hungry) - A rule which causes \texttt{hungry} to be true is required.
\begin{lstlisting}[basicstyle=\scriptsize, escapechar=^]
cross(V) causes hungry if numCrossing(V, strongMissionary) = 0.
\end{lstlisting}

MCP \#17 (cannibals can become hungry) - A rule which causes \texttt{hungry} to be true is required.
\begin{lstlisting}[basicstyle=\scriptsize, escapechar=^]
cross(V) causes hungry if numCrossing(V, strongMissionary) = 0.
\end{lstlisting}

MCP \#10 (a missionary can walk on water) - Rules missing which represent ``It cannot be that Jesus is on the boat, but the number of Jesus’s on the boat is not 1, or that Jesus is not on the boat and the number of Jesus’s is not 0.''
\begin{lstlisting}[basicstyle=\scriptsize, escapechar=^]
nonexecutable cross(Vessel) if jesusOnBoat(Vessel) & loading(Vessel, jesus) \= 1.
nonexecutable cross(Vessel) if ~jesusOnBoat(Vessel) & loading(Vessel, jesus) \=0.
\end{lstlisting}

MCP \#10 (a missionary can walk on water) - Updating Jesus's location and count on each side is missing.
\begin{lstlisting}[basicstyle=\scriptsize, escapechar=^]
cross(Vessel) causes jesusLocation = Loc if jesusOnBoat(Vessel) & jesusLocation = Loc1 & Loc\=Loc1.
walk increments numOnBank(jesus, Loc) by 1 if jesusLocation = Loc1 & Loc\=Loc1.
walk decrements numOnBank(jesus, Loc) by 1 if jesusLocation = Loc.
\end{lstlisting}

\noindent
\textbf{Harmful rules which represent something not specified in the problem.}
MCP \# 6 (only one missionary and one cannibal can row) - The following rule disallows both the missionary rower and the cannibal rower to be on the boat at the same time. This condition makes it impossible to cross all missionaries and cannibals. Thus it should simply be removed.
\begin{lstlisting}[basicstyle=\scriptsize, escapechar=^]
nonexecutable cross(boat) if rower(boat, missionaries) & rower(boat, cannibals).
\end{lstlisting}

MCP \# 6 (only one missionary and one cannibal can row) - There is no reasno to disallow botht he missionary and cannibal rower to be on board. The following should be removed.
\begin{lstlisting}[basicstyle=\scriptsize, escapechar=^]
% 13. The crossing action is not executable if both missionaries and cannibals are rowing.
nonexecutable cross(boat) if rower(boat, missionaries) & rower(boat, cannibals).
\end{lstlisting}

Sudoku (var1) - the following default rule is incorrect. It allows the initial values to be changed in a later step, and changing of the board from unsatisfiable to satisfiable.
\begin{lstlisting}[basicstyle=\scriptsize, escapechar=^]
% Default value of val is none
default val(R, C) = none.
\end{lstlisting}

MCP \#5 (an oar on each bank) - The following constraints are incorrect because the variables N1, N2, and N3 are quantified over all integer values. It should instead be represented with an implication (always (…) → (…). What is trying to be represented (the natural language writing) is not necessary. Therefore they can be simply removed.
\begin{lstlisting}[basicstyle=\scriptsize, escapechar=^]
always N1 + N2 + N3 = 3 & N1 = numMissionariesAt(bank1) & N2 = numMissionariesAt(bank2) & N3 = numMissionariesOn(boat).
% 11. Total number of cannibals is always 3.
always N1 + N2 + N3 = 3 & N1 = numCannibalsAt(bank1) & N2 = numCannibalsAt(bank2) & N3 = numCannibalsOn(boat).
% 12. Total number of oars is always 2.
always N1 + N2 + N3 = 2 & N1 = numOarsAt(bank1) & N2 = numOarsAt(bank2) & N3 = numOarsOn(boat).
\end{lstlisting}

MCP \#10 (a missionary can walk on water) - The following makes it so the boat’s location and Jesus’s location cannot change. They should only be set in the query at time 0. These should be removed.
\begin{lstlisting}[basicstyle=\scriptsize, escapechar=^]
% Initial conditions
boatLocation = bank1.
jesusLocation = bank1.
\end{lstlisting}

name - explanation
\begin{lstlisting}[basicstyle=\scriptsize, escapechar=^]

\end{lstlisting}

\noindent
\textbf{Harmful rules attempting to represent an aspect of the problem.}
MCP \# 17 (cannibals can become hungry) - The following constraint about cannibals outnumbering missionaries should be changed to include the condition \texttt{hungry}. This is because the problem description states that if the cannibals are not hungry, then the missionaries will not be eaten.

\begin{lstlisting}[basicstyle=\scriptsize, escapechar=^]
impossible numAt(missionaries, Loc) = N1 &
numAt(strongMissionary, Loc) = N2 &
numAt(cannibals, Loc) = N3 &
N1 + N2 > 0 &
N3 > N1 + N2 ^{\cred \& hungry}^. ^{\cred % hungry should be added}^
\end{lstlisting}

MCP \#2 (missionaries and cannibals can exchange hats) - Incrementing/decrementing does not work properly due to $\bcp$ semantics. 
\begin{lstlisting}[basicstyle=\scriptsize, escapechar=^]
% For missionaries:
cross(V) decrements numMissionariesAt(bank1) by 1 if onboard(V, Ms) & loc_member(Ms) = bank1 & loc_vessel(V) = bank1.
cross(V) increments numMissionariesAt(bank2) by 1 if onboard(V, Ms) & loc_member(Ms) = bank1 & loc_vessel(V) = bank1.

cross(V) decrements numMissionariesAt(bank2) by 1 if onboard(V, Ms) & loc_member(Ms) = bank2 & loc_vessel(V) = bank2.
cross(V) increments numMissionariesAt(bank1) by 1 if onboard(V, Ms) & loc_member(Ms) = bank2 & loc_vessel(V) = bank2.

% For cannibals:
cross(V) decrements numCannibalsAt(bank1) by 1 if onboard(V, Cs) & loc_member(Cs) = bank1 & loc_vessel(V) = bank1.
cross(V) increments numCannibalsAt(bank2) by 1 if onboard(V, Cs) & loc_member(Cs) = bank1 & loc_vessel(V) = bank1.

cross(V) decrements numCannibalsAt(bank2) by 1 if onboard(V, Cs) & loc_member(Cs) = bank2 & loc_vessel(V) = bank2.
cross(V) increments numCannibalsAt(bank1) by 1 if onboard(V, Cs) & loc_member(Cs) = bank2 & loc_vessel(V) = bank2.
\end{lstlisting}

These should be corrected to:
\begin{lstlisting}[basicstyle=\scriptsize, escapechar=^]
cross(V) decrements numMissionariesAt(L) by N if numMissionariesOnBoard(V) = N & loc_vessel(V) = L.
cross(V) increments numMissionariesAt(L) by N if numMissionariesOnBoard(V) = N & loc_vessel(V) = L1 & L\=L1.

cross(V) decrements numCannibalsAt(L) by N if numCannibalsOnBoard(V) = N & loc_vessel(V) = L.
cross(V) increments numCannibalsAt(L) by N if numCannibalsOnBoard(V) = N & loc_vessel(V) = L1 & L\=L1.
\end{lstlisting}

MCP \#9 (big cannibal and small missionary) - The disjunction should be encapsulated in parentheses.
\begin{lstlisting}[basicstyle=\scriptsize, escapechar=^]
nonexecutable cross2(P1, P2) if ^{\cred loc(P1) \textbackslash= loc(boat) $\vert$ loc(P2) \textbackslash = loc(boat). \% The disjunction should be encapsulated in parentheses}
\end{lstlisting}

MCP \#9 (big cannibal and small missionary) - The following laws need to assure that the location where the missionaries/cannibals are incremented is not where the boat is.

\begin{lstlisting}[basicstyle=\scriptsize, escapechar=^]
cross1(P) increments num_missionaries(NewLoc) by 1 if P = M & NewLoc \= Loc. 
cross1(P) increments num_cannibals(NewLoc) by 1 if P = C & NewLoc \= Loc.
\end{lstlisting}
These should be corrected to:
\begin{lstlisting}[basicstyle=\scriptsize, escapechar=^]
cross1(P) increments num_missionaries(NewLoc) by 1 if P = M & NewLoc \= Loc & loc(boat) = Loc. 
cross1(P) increments num_cannibals(NewLoc) by 1 if P = C & NewLoc \= Loc & loc(boat) = Loc.
\end{lstlisting}

MCP \#9 (big cannibal and small missionary) - The term ``else'' is not recognized in {\sc cplus2asp}.
\begin{lstlisting}[basicstyle=\scriptsize, escapechar=^]
cross2(P1, P2) decrements num_missionaries(Loc) by L1 if L1 = N1 + N2 & N1 = 1 if P1 = M ^{\cred  else}^ N1 = 0 & N2 = 1 if P2 = M ^{\cred else}^ N2 = 0 & loc(P1) = Loc & loc(P2) = Loc. 
cross2(P1, P2) decrements num_cannibals(Loc) by L2 if L2 = N1 + N2 & N1 = 1 if P1 = C ^{\cred else}^ N1 = 0 & N2 = 1 if P2 = C ^{\cred else}^ N2 = 0 & loc(P1) = Loc & loc(P2) = Loc. 
cross2(P1, P2) increments num_missionaries(NewLoc) by L1 if L1 = N1 + N2 & N1 = 1 if P1 = M ^{\cred else}^ N1 = 0 & N2 = 1 if P2 = M ^{\cred else}^ N2 = 0 & NewLoc \= Loc & loc(P1) = Loc & loc(P2) = Loc. 
cross2(P1, P2) increments num_cannibals(NewLoc) by L2 if L2 = N1 + N2 & N1 = 1 if P1 = C ^{\cred else}^ N1 = 0 & N2 = 1 if P2 = C ^{\cred else}^ N2 = 0 & NewLoc \= Loc & loc(P1) = Loc & loc(P2) = Loc.
\end{lstlisting}
These should be corrected to:

\begin{lstlisting}[basicstyle=\scriptsize, escapechar=^]
cross2(M1, M2) decrements num_missionaries(Loc) by 2 if loc(boat) = Loc.
cross2(C1, C2) decrements num_cannibals(Loc) by 2 if loc(boat) = Loc.
cross2(M1, C1) decrements num_missionaries(Loc) by 1 if loc(boat) = Loc.
cross2(C1, M1) decrements num_missionaries(Loc) by 1 if loc(boat) = Loc.
cross2(M1, C1) decrements num_cannibals(Loc) by 1 if loc(boat) = Loc.
cross2(C1, M1) decrements num_cannibals(Loc) by 1 if loc(boat) = Loc.

cross2(M1, M2) increments num_missionaries(Loc) by 2 if loc(boat) = Loc1 & Loc\=Loc1.
cross2(C1, C2) increments num_cannibals(Loc) by 2 if loc(boat) = Loc1 & Loc\=Loc1.
cross2(M1, C1) increments num_missionaries(Loc) by 1 if loc(boat) = Loc1 & Loc\=Loc1.
cross2(C1, M1) increments num_missionaries(Loc) by 1 if loc(boat) = Loc1 & Loc\=Loc1.
cross2(M1, C1) increments num_cannibals(Loc) by 1 if loc(boat) = Loc1 & Loc\=Loc1.
cross2(C1, M1) increments num_cannibals(Loc) by 1 if loc(boat) = Loc1 & Loc\=Loc1.
\end{lstlisting}

MCP \#10 (a missionary can walk on water) - This rules incorrectly disallows crossing if there are more cannibals than missionaries on board, but it should only disallow it if there are more \textit{and} there is at least one missionary on board.
\begin{lstlisting}[basicstyle=\scriptsize, escapechar=^]
% 9b. Crossing the boat is not executable if cannibals loaded onto the boat exceed missionaries loaded onto the boat if jesusOnBoat(Vessel) = false.
nonexecutable cross(Vessel) if loading(Vessel, cannibals) = N_c & loading(Vessel, missionaries) = N_m & jesusOnBoat(Vessel) = false & N_c > N_m.
^{\cred \% N\_m $>$ 0 should be added to the if clause }^
\end{lstlisting}

MCP \#10 (a missionary can walk on water) - Similarly, there being at least one missionary needs to be adding to the body of these rules to make them correct.
\begin{lstlisting}[basicstyle=\scriptsize, escapechar=^]
% 10b. Crossing the boat is not executable if cannibals remaining on the starting bank after loading would outnumber missionaries remaining on the starting bank when Jesus is not at that bank.
nonexecutable cross(Vessel) if boatLocation = Loc & jesusLocation \= Loc &
numOnBank(cannibals, Loc) = N_c & loading(Vessel, cannibals) = L_c & Rem_c = N_c - L_c &
numOnBank(missionaries, Loc) = N_m & loading(Vessel, missionaries) = L_m & Rem_m = N_m - L_m &
Rem_c > Rem_m.
^{\cred \% Rem\_m $>$ 0, needs to be added to the if clause}^
\end{lstlisting}

MCP \#10 (a missionary can walk on water) - Similarly, there being at least one missionary needs to be adding to the body of this rule to make it correct.
\begin{lstlisting}[basicstyle=\scriptsize, escapechar=^]
It should be possible for the number of cannibals to be larger than the number of missionaries if there are no missionaries on the bank.
generated:
% 14b. It is impossible for cannibals to outnumber missionaries on any bank when Jesus is not at that bank.
impossible N_c > N_m & numOnBank(cannibals, Loc) = N_c & numOnBank(missionaries, Loc) = N_m & jesusLocation \= Loc.
^{\cred \% N\_m $>$ 0, needs to be added to the if clause}^
\end{lstlisting}

\subsection{Query}

\textbf{Syntax issues.}
Sudoku1 - The following query is written with incorrect keywords ``initially'' and ``goal'' to represent the initial and goal state of Sudoku.

\begin{lstlisting}[basicstyle=\scriptsize, escapechar=^]
^{\cred initially}^ valueAt(0,0)=8.
^{\cred initially}^ valueAt(1,2)=3.
^{\cred initially}^ valueAt(1,3)=6...
^{\cred goal}^ valueAt(R,C)\=0.
\end{lstlisting}

Sudoku (var2)  - ``maxstep: 81'' is not syntactically correct.
\begin{lstlisting}[basicstyle=\scriptsize, escapechar=^]
:- query
    0:
        val(1,2)=9 & val(1,4)=3 & val(1,9)=1 &
        val(2,5)=8 & val(2,8)=4 & val(2,9)=6 &
        val(3,7)=8 &
        val(4,1)=4 & val(4,3)=5 & val(4,5)=6 & val(4,8)=3 &
        val(5,3)=3 & val(5,4)=2 & val(5,5)=7 & val(5,6)=5 & val(5,7)=6 &
        val(6,2)=6 & val(6,5)=1 & val(6,7)=9 & val(6,9)=4 &
        val(7,3)=1 &
        val(8,1)=5 & val(8,2)=8 & val(8,5)=2 &
        val(9,1)=2 & val(9,6)=7 & val(9,8)=6;
    ^{\cred maxstep: 81. \% should be ``maxstep :: 81.'', alternatively it can be removed)}^
\end{lstlisting}

\noindent
\textbf{Semantic issues.}
Tower of Hanoi (7, var1) - The following query goal condition incorrectly states that that every disk should be on peg c. Instead, they should be stacked in order, with only disk \texttt{d1} on c.

\begin{lstlisting}[basicstyle=\scriptsize, escapechar=^]
:- query
    0: loc(d1)=c & loc(d2)=d1 & loc(d3)=d2 & loc(d4)=a & loc(d5)=b & loc(d6)=d4 & loc(d7)=d5 &
       weight(d1)=7 & weight(d2)=6 & weight(d3)=5 & weight(d4)=4 & weight(d5)=3 & weight(d6)=2 & weight(d7)=1;
    maxstep: loc(d1)=c & loc(d2)=^{\cred c}^ & loc(d3)=^{\cred c}^ & loc(d4)=^{\cred c}^ & loc(d5)=^{\cred c}^ & loc(d6)=^{\cred c}^ & loc(d7)=^{\cred c}^.

^{\cred \% The maxstep should be as follows:}^
^{\cred maxstep: loc(d1)=c \& loc(d2)=d1 \& loc(d3)=d2 \& loc(d4)=d3 \& }^ 
^{\cred loc(d5)=d4 \& loc(d6)=d5 \& loc(d7)=d6.}^
\end{lstlisting}

MCP \#3 (there are 4 missionaries and 4 cannibals) - There are two issues: (1) The query includes a \texttt{maxstep} of 11, which it should not, (2) the query uses 3 for the number of missionaries and cannibals, this should be 4.
\begin{lstlisting}[basicstyle=\scriptsize, escapechar=^]
:- query
    0: numOn(bank1, missionaries) = ^{\cred 3}^ & numOn(bank1, cannibals) = ^{\cred 3}^ & loc(boat) = bank1 & numOn(bank2, missionaries) = 0 & numOn(bank2, cannibals) = 0;
    ^{\cred maxstep :: 11; \% (should be removed)}^
    maxstep: numOn(bank2, missionaries) = ^{\cred 3}^ & numOn(bank2, cannibals) = ^{\cred 3}^.
\end{lstlisting}

\section{Missionaries and Cannibals Elaborations} \label{sec:mcp_elaborations}

In \citet{mccarthy98elaboration}, there are 20 elaborations discussed in which we include 16 in this work, as well as the original problem:

``Three missionaries and three cannibals come to a river
and find a boat that holds two. If the cannibals ever outnumber the missionaries on either bank, the missionaries
will be eaten.
How shall they cross?''

\begin{itemize}
    \item []\#1 ``The boat is a rowboat. (Or the boat is a motorboat). By itself this
is a trivial elaboration. Adding it should not affect the reasoning. By
default, a tool, i.e. the boat, is usable. Further elaborations might use
specific properties of rowboats.''
\item []\#2``The missionaries and cannibals have hats, all different—another trivial
elaboration. These hats may be exchanged among the missionaries and
cannibals.''
\item []\#3``There are four missionaries and four cannibals. The problem is now
unsolvable.''
\item []\#4 - ``The boat can carry three.'' Five missionaries and cannibals can cross, but not six. We use the version with six members of each group, which no solution is possible.
\item []\#5 - ``There is an oar on each bank. One person can cross in the boat with
just one oar, but two oars are needed if the boat is to carry two people.''
\item []\#6 - ``Only one missionary and one cannibal can row.''
\item []\#7 - ``The missionaries can’t row. This makes the problem impossible, since
any solution requires two missionaries in the boat at some time''
\item []\#8 - ``The biggest cannibal cannot fit in the boat with another person.''
\item []\#9 - ``If the biggest cannibal is isolated with the smallest missionary, the latter will be eaten.''
\item []\#10 - ``One of the missionaries is Jesus Christ.'' This makes it so that one of the missionaries can walk on water, which has a solution with only 7 steps.
\item []\#11 - ``Three missionaries alone with a cannibal can convert him into a missionary.''
\item []\#13 - ``There is a bridge. '' This makes makes it so any number can cross, we use the version with 5 missionaries and cannibals.
\item []\#14 - ``The boat leaks and must be bailed concurrently with rowing.'' We use the version from \citet{Lifschitz00MissionariesAC} which assumes that bailing is only needed when there are two people in the boat, and that one person could not row and bail at the same time.
\item []\#16 - ``There is an island. '' This makes it so any number can cross. We use an instance with 5 missionaries and cannibals.
\item []\#17 - ``There are four cannibals and four missionaries, but if the strongest of
the missionaries rows fast enough, the cannibals won’t have gotten so
hungry that they will eat the missionaries.'' This has some vagueness, we make it more concrete by specifying that as long as the strongest missionary rows then the cannibals will not be hungry.
\item []\#19 - ``There are two sets of missionaries and cannibals too far apart along
the river to interact. The two problem should be solvable separately
without considering interleaving actions at the two sites.'' Here we assume that the two sets are two identical versions of the original problem.
\end{itemize}

\end{document}